\documentclass[10pt,twocolumn,letterpaper]{article}

\usepackage[pagenumbers]{cvpr} 

\usepackage{cuted}
\usepackage{multirow, tabularx}
\usepackage{subcaption}
\usepackage[table]{xcolor}
\usepackage[detect-all,per-mode=symbol]{siunitx}
\usepackage{amssymb}
\usepackage[T1]{fontenc}

\newcommand{\mb}{\mathbf}
\newcommand{\cmark}{\checkmark}%
\newcommand{\xmark}{-}%

\definecolor{cvprblue}{rgb}{0.21,0.49,0.74}
\usepackage[pagebackref,breaklinks,colorlinks,allcolors=cvprblue]{hyperref}

\def\paperID{215} 
\def\confName{3DV\xspace}
\def\confYear{2027\xspace}

\title{CQF-HMR: Continuous Quaternion Flows for \\ Probabilistic 3D Human Mesh Recovery from a Single Image}

\author{
\small{Cuong Le\textsuperscript{1}, Bao-Long Tran\textsuperscript{1}, Pavlo Melnyk\textsuperscript{1}, Tahereh Dehdarirad\textsuperscript{1}, Bastian Wandt\textsuperscript{2}, Mårten Wadenbäck\textsuperscript{1}} \\
\small{\textsuperscript{1} Linköping University, \textsuperscript{2} Independent researcher}. Email: \{\textit{firstname}\}.\{\textit{lastname}\}@liu.se \\
}

\begin{document}
\maketitle

\begin{strip}
    \centering
    \captionsetup{type=figure}
    \includegraphics[width=0.99\textwidth]{figures/Teaser.png}
    \captionof{figure}{Conditioned on a 2D input estimated from a single image, our CQF-HMR maps a randomly sampled 3D human pose to a plausible one using the continuous normalizing flows under the quaternion constraint. Unlike other 3D rotation representations, quaternion flows provide smoother generation trajectories, leading to a more plausible 3D human mesh recovery.}
    \label{fig:teaser}
\end{strip}

\begin{abstract}
Recovering 3D digital humans from a single 2D image is an ill-posed computer vision problem due to the loss of depth information.
Probabilistic 3D human pose estimation compensates for this by estimating a set of 3D hypotheses from a prior distribution via generative models.
However, most prior work focuses only on 3D keypoints, which often leads to implausible poses that are difficult to apply to downstream tasks, \eg animation or digital humans.
SMPL-based methods are more scalable thanks to the explicit body priors, but it requires more complex modeling of the generation process due to the non-additive nature of the joint rotations.
In this work, we propose a novel approach for probabilistic 3D humans using quaternion-constrained continuous normalizing flows conditioned on 2D pose estimations.
Our proposed quaternion flows show significant advantages over approaches using other rotation representations. 
Experiments demonstrate state-of-the-art results of our method on Human3.6M, particularly in ambiguous settings, and comparable pose estimation accuracy on challenging 3DPW and EMDB benchmarks.
\end{abstract}    
\section{Introduction}
\label{sec:intro}

The recovery of 3D human pose and shape from a single view remains a challenge in computer vision due to the depth ambiguity of the 2D input image.
Multiple valid yet diverse 3D poses can project to the same 2D observation, resulting in a highly ill-pose problem.
Therefore, recent work in single-view 3D human pose estimation (HPE) steers the focus to the estimation of the full 3D distribution or a set of pose hypotheses using generative models conditioned on 2D inputs~\cite{jahangiri_iccvw17,li_cvpr2022,wehrbein_iccv21,holmquist_iccv23,wehrbein_wacv25,le_iclr26}.
While traditional 3D-keypoint methods achieve lower errors on the training data, they often produce unnatural poses when encountering in-the-wild samples~\cite{bogo_eccv16}.
On the other hand, SMPL-based approaches, often called human mesh recovery (HMR), address the issue via explicit body kinematics constraints imposed by the SMPL mesh model~\cite{loper_acm15}.
Moreover, SMPL-based predictions are easily applied to multiple applications, such as AR/VR, simulation or biomechanics, making them the preferable approach for 3D HPE~\cite{rempe_humor21,zhao_cvpr26}.

Probabilistic 3D HPE approaches learn the mapping trajectory from a prior distribution, \eg the Gaussian distribution, to the ground-truth 3D human pose distribution collected from motion capture systems.
Previous work samples the set of 3D keypoints from the Gaussian distribution and learn the vector field to move these keypoints to the true 3D human body configuration based on the 2D condition, often via diffusion or flow-based models~\cite{wehrbein_iccv21,holmquist_iccv23,le_tmlr26}.
The same idea is also applied to SMPL-based methods, by sampling random joint 3D rotations, often in axis-angles and map them to the ground-truth data~\cite{zanfir_eccv20,aliakbarian_cvpr22,sfikas_bmvcw25}.
However, due to the non-additive nature of 3D rotations, applying vanilla flow-based generative models on SMPL's joint rotations can introduce discontinuities and results in wrong 3D poses during the generation process~\cite{le_iclr26}.
On the other hand, the unit quaternion, with only one additional parameter, is free of singularities and its differential equation is globally defined, ensuring a smooth rotating trajectory for the generation of 3D rotations.
To this end, we propose a novel framework for probabilistic 3D HMR using continuous quaternion normalizing flows, CQF-HMR, for learning the generation of SMPL 3D joint rotations conditioned on 2D inputs.

The underlying differential equation of CQF-HMR is learned via the flow matching framework~\cite{lipman_iclr23}, in which the optimal transport is designed to be the spherical linear interpolation between the randomly sampled quaternion and ground-truth.
Because of the non-additive nature of 3D rotations, we formulate the quaternion integration for solving the differential equation via an operation based on the Hamilton product. 
This operation ensures the unit quaternion stays on the spherical constrain of $\mathcal{S}^{3}$ without the need for re-normalization as other representations~\cite{le_iclr26,golabek_aircraft22}.
The SMPL shape parameter is learned via the flow matching with straight OT path.
The condition embedding is learned by a graph neural network on the estimated 2D heatmaps extracted from a 2D detector HRNet~\cite{sun_cvpr19}.
Beside the common Gaussian distribution as prior which could create highly implausible poses, we also investigate the learning from a more sophisticated human prior, VPoser~\cite{pavlakos_cvpr19}, for sampling plausible initial poses.
The proposed method is evaluated against related probabilistic approaches on the Human3.6M~\cite{ionescu_pami14}, 3DPW~\cite{marcard_eccv18} and EMDB~\cite{kaufmann_iccv23} datasets.
In summary, our main contributions are:
\begin{itemize}
    \item We show that using quaternion flows with a unit-sphere constraint produces smooth, continuous generative trajectories for probabilistic 3D human mesh recovery.
    \item We integrate the design of quaternion differential equation of SMPL body joints into the flow matching pipeline, utilizing 2D heatmaps as a condition.
    \item We set a new state-of-the-art result on the Human3.6M benchmark, including its ambiguous split.
    
\end{itemize}
\section{Related work}
\label{sec:related_work}

\subsection{Monocular 3D pose and mesh estimation}
\label{subsec:related_3dhpe}
The field of monocular 3D HPE consists of two main directions: 1) direct estimation from images~\citep{mehta_vnect17,rogez_tpami19}, and 2) lifting from 2D cues~\citep{martinez_iccv17,yeh_nips19,liu_cvpr20}.
Due to the maturity of the 2D HPE, the 2D-3D lifting is the more popular approach and often produce more accurate 3D estimations.
The famous work from~\cite{pavllo_cvpr19} introduces the use of multiple 2D pose as input to leverage the temporal information for more accurate 3D HPE.
Many follow-up studies use the same approach with better learning models, \eg transformers~\cite{zheng_iccv21,zhao_cvpr23,peng_cvpr24}.
However, these methods require synchronized video inputs and is not applicable to single-image captures that often occur in-the-wild.

The single-image approach is more suitable for in-the-wild examples.
The pioneer work from~\cite{martinez_iccv17} predicts 3D human pose from a single 2D pose using a two-layer ResNet architecture.
Using graph networks is also a popular approach by modeling the natural human body connection between joints~\cite{cai_iccv19}.
The traditional 2D-3D lifting methods predict only 3D keypoints and bypass the human body constraints, \ie bone-length plausibility along the camera depth axis.
To impose the natural body constraints, recent work uses the SMPL mesh model as the prior~\cite{loper_acm15} and only estimate the joint rotations based on the 2D/3D cues~\cite{kolotouros_iccv19,kocabas_cvpr20}.
\citet{goel_iccv23} predict the SMPL parameters from a single image using the transformer with the cross-attention module.
\citet{li_hybrik21} utilize the highly accurate 3D keypoint estimation model and solve the analytical inverse kinematics to recover plausible twist and swing angles for the SMPL model.
A major issue of single-image approaches is their single prediction of the most likely 3D human pose, which can be overly confident and incorrect.

\subsection{Probabilistic 3D pose and shape estimation}
\label{subsec:related_prob}
Recovering 3D from 2D is ill-posed because many 3D poses can project to the same 2D cue, thus requiring a full modeling of 3D pose distribution.
The emerging field of probabilistic 3D HPE address the challenge via generative models that produce multiple 3D hypotheses for the approximation of the posterior distribution.
\citet{wehrbein_iccv21} use the normalizing flows conditioned on the 2D posed, extracted from HRNet~\cite{sun_cvpr19}, to map the source Gaussian distribution to the target 3D pose distribution.
Similarly, DiffPose~\cite{holmquist_iccv23} learns the distribution mapping via the diffusion model, using the same 2D conditions.
The diffusion models follow the stochastic trajectory, which results in sub-optimal mapping for the estimation of 3D pose hypotheses.
Recent work~\cite{wang_cvpr26,le_tmlr26} instead learns the optimal-transport straight trajectory via the flow matching framework and achieve more accurate 3D poses.
To enforce the body constraints, \citet{kolotouros_iccv21} recovers the target distribution of the SMPL's pose and shape parameters via normalizing flows.
Also with the normalizing flows, \citet{wehrbein_wacv25} utilizes the uncertainty from the 2D estimator heatmaps for more accurate 3D mesh estimation and penalize incorrect hypotheses via human segmentation masks.

\begin{figure*}[t]
    \centering
    \includegraphics[width=0.98\linewidth]{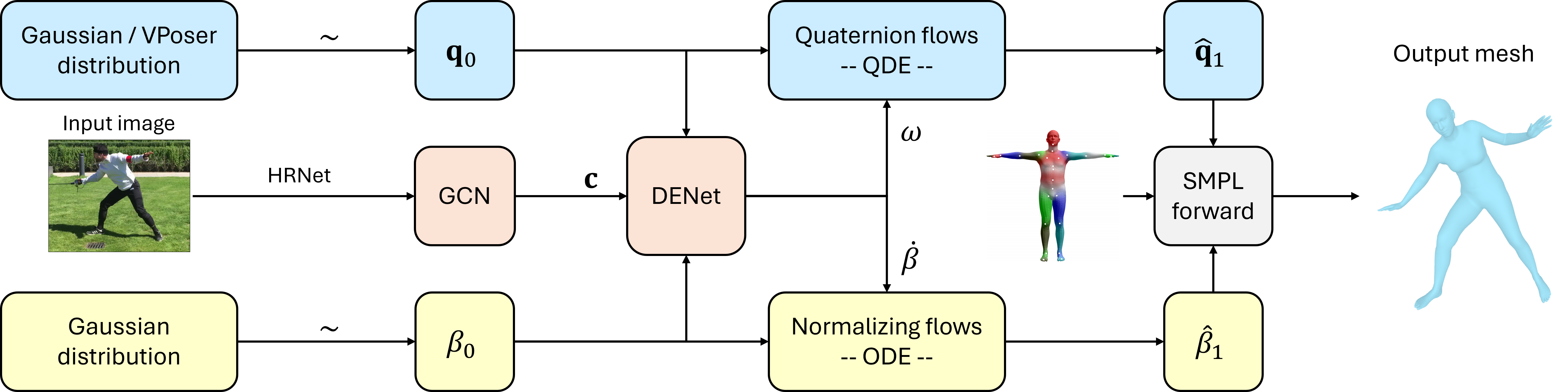}
    \caption{
    Pipeline of CQF-HMR.
    The method consists of two differential equations: an QDE for the quaternion flows and an ODE for the shape normalizing flows.
    The initial pose $\mb{q}_0$ and shape $\beta_0$ are sampled from source distributions, \ie Gaussian distribution, while GCN encodes the 2D heatmaps extracted by HRNet to obtain the lifting condition $\mb{c}$.
    DENet estimates the angular velocity $\omega$ and shape shifting rate $\dot{\beta}$ for the predictions of quaternion pose $\hat{\mb{q}}_1$ and shape $\hat{\beta}_1$ under the $\mathcal{S}^3$ constraint with OT transports.
    The 3D mesh is obtained by passing the predicted pose and shape to the SMPL transformation layer.
    The pipeline is executed in parallel for multiple hypotheses.
    }
    \label{fig:pipeline}
\end{figure*}

\section{Method}
\label{sec:method}

We formulate the continuous normalizing flows (CNF) as a quaternion-based kinematics process and aim to recover the human 3D joint rotations based on 2D inputs.
We first apply an off-the-shelf detector, HRNet~\cite{sun_cvpr19}, to extract the probability heatmaps of 2D joints and use these information as the condition for the generative process.
With $\mb{q}\in\mathbb{R}^{4}$ denoting a quaternion that represents the 3D pose of a human joint, we learn the optimal transport angular velocity $\omega\in\mathbb{R}^3$ to map the pose $q$ sampled from a prior distribution to the ground-truth pose, and the trajectory traverses along the unit-sphere constraint $\mathcal{S}^{3}$ of quaternions.

\subsection{Quaternion differential equation}
\label{subsec:method_qde}

In the Hamilton set $\mathbb{H}$, a quaternion $\mb{q}\in\mathbb{R}^{4}$ is represented by a 4D vector $(\textup{w}, v_1, v_2, v_3)$, also written as:
\begin{equation}
    \label{eq:quaternion}
    \mb{q} = \textup{w} + v_1i + v_2j + v_3k = (\textup{w}, \mb{v})~,
\end{equation}
where $i,j,k$ are imaginary units satisfying $i^2 = j^2 = k^2 = ijk = -1$.
To correctly represent 3D rotations, the quaternion is strictly unit, \ie normalized to unit length, $\lVert \mb{q} \rVert = 1$.
We refer to all quaternions in this paper as unit quaternions for convenience.
The Hamilton product combines two quaternions, $\mb{q}=(\textup{w}_1,\mb{v}_1)$ and $\mb{q}_2=(\textup{w}_2,\mb{v}_2)$, into a quaternion representing a combined pose, and is defined as $\mb{q}_1 \otimes \mb{q}_2 = \textup{w}_1\textup{w}_2 - \mb{v}_1^{\top} \mb{v}_2, \textup{w}_2\mb{v}_1 + \textup{w}_1\mb{v}_2 + \mb{v}_1 \times \mb{v}_2$, with the crucial non-commutativity property of $\mb{q}_1 \otimes \mb{q}_2 \neq \mb{q}_2 \otimes \mb{q}_1$.

The quaternion flow is characterized by an ordinary differential equation (ODE) of the quaternion-valued variables, \ie quaternion differential equation (QDE)~\cite{le_tmlr26,yang_spacecraft19,kuipers_quaternions99}, defined as:
\begin{equation}
    \label{eq:qde}
    \dot{\mb{q}} = \frac{1}{2}
    \begin{bmatrix}
        -[\omega]_\times & \omega \\ -\omega^\top & 0 \\
    \end{bmatrix}
    \mb{q}, \quad
    [\mb{\omega]}_\times =
    \begin{bmatrix}
        0 & -\omega_3 & \omega_2 \\
        \omega_3 & 0 & -\omega_1 \\
        -\omega_2 & \omega_1 & 0
    \end{bmatrix},
\end{equation}
where $\omega\in\mathbb{R}^3$ is the angular velocity represented as a scaled axis vector.
Unlike the vanilla ODE solvers~\cite{chen_nips18,vuik_numericalode23}, the Hamilton product is required to solve the QDE numerically, since we require that the solution abide by the $\mathcal{S}^3$ constraint, \ie be of unit length, which is not preserved by addition.
We implement the quaternion integration with as:
\begin{equation}
    \label{eq:solution_qde}
    \begin{split}
        \mb{q}_{t+\Delta t} & = \mb{q}_{t} \otimes \textup{quat}_{\Delta t}(\omega_{t+\Delta t}) \\
        & = \mb{q}_t \otimes \textstyle\left(\cos{\left(\frac{\theta}{2}\right)},~\sin{\left(\frac{\theta}{2}\right)}\frac{\omega_t}{\lVert \omega_t\rVert}\right)~,
    \end{split}
\end{equation}
where $\theta=\lVert\omega_t\rVert \Delta t$ is the scalar magnitude of the rotation scaled by a time step $\Delta t$, ${\omega}_{t}$ is normalized to a unit vector, and $\textup{quat}_{\Delta t}(\omega_{t+\Delta t})$ is the quaternionized angular velocity.
This integration scheme guarantees the unit-sphere $\mathcal{S}^3$ requirement of unit quaternions, resulting in smooth and continuous generation trajectories of 3D joint rotations.

\subsection{Quaternion flow matching}
\label{subsec:method_qfm}
Consider the mapping bounded in $t=[0,1]$ from the pose $\mb{q}_0$ sampled from a prior distribution to the 3D joint pose $\mb{q}_1$, the diffeomorphic flow from $\mb{q}_0$ to $\mb{q}_1$ is defined via the QDE in~\cref{eq:solution_qde}, and the angular velocity is estimated as:
\begin{equation}
    \label{eq:para_qde}
    \omega_t = f_{\textup{QDE}}(\mb{q}_t,t,\mb{c})~,
\end{equation}
where $f_{\textup{QDE}}(\mb{q}_t,t,\mb{c})$ is a parameterized multi-layer perceptrons (MLP), $t\in[0,1]$, and $\mb{c}\in\mathbb{R}^{d}$ is the $d$-dimensional conditioning vector extracted from 2D inputs.

Unlike the traditional CNF frameworks that learn the unstable normalizing flows by maximizing the likelihood of $\mb{q}_1$, the flow matching framework from~\cite{lipman_iclr23} leverages the conditional flows between $\mb{q}_0$ and $\mb{q}_1$ via the optimal transport trajectory, resulting in a more stable and efficient learning process.
As inspiration~\cite{yue_icml25}, we replace the vanilla OT flow matching path of linear interpolation (\texttt{lerp}) to the spherical linear interpolation (\texttt{slerp}) as the OT path for quaternions.
The \texttt{slerp} between $\mb{q}_0$ and $\mb{q}_1$ is defined as:
\begin{equation}
    \label{eq:slerp}
    \begin{split}
    & \mb{q}_t = \textup{\texttt{slerp}}(\mb{q}_0,\mb{q}_1,t) = \mb{q}_0(\mb{q}_0^{-1}{\mb{q}}_1)^{t}~, \\
    & \Rightarrow \dot{\mb{q}}^* = \log(\mb{q}_0^{-1}\mb{q}_1)~,
    \end{split}
\end{equation}
where $\dot{\mb{q}}^*$ is the OT quaternion derivative, and the corresponding angular velocity $\omega^*$ can be approximated by only taking the vector part of $\dot{\mb{q}}^*$.
Note that, \texttt{slerp} is a linear interpolation with a constant derivative, thus the OT angular velocity $\omega^*$ is independent of time.
The learning objective is a simple regression task minimizing the loss:
\begin{equation}
    \mathcal{L_\omega} = \lVert f_{\textup{QDE}}(\mb{q}_t,t,\mb{c}) - \omega^* \rVert _2^2~, \, \forall t \in [0, 1]~.
\end{equation}

To instantiate the pose $\mb{q}_0$, we sample an axis-angle pose $\theta_0\in\mathbb{R}^{3}$ from the source distribution, \eg Gaussian distribution $\theta_0\sim\mathcal{N}(0,\mb{I})$, and convert the pose to quaternion.
Due to the kinematics chain constraint of the human body, \ie parent pose affects the child pose, sampling from Gaussian distribution introduces highly implausible initial poses.
To address this issue, we alternatively sample $\mb{q}_0$ from the human body prior $\mathcal{D}$ (VPoser~\cite{pavlakos_cvpr19}), expressed as:
\begin{equation}
    \label{eq:vposer}
    \theta = \mathcal{D}(z)~\mid \mb{z} \sim \mathcal{N}(0,\mb{I})~, \\
\end{equation}
where $\mb{z}\in\mathbb{R}^{32}$ is the VPoser latent embedding sampled from Gaussian distribution.

\subsection{2D conditioning}
\label{subsec:method_condition}

Similar to~\cite{holmquist_iccv23,le_tmlr26}, we pre-extract a set of 2D heatmaps for $J$ joints in MPII format using the HRNet detector ~\cite{sun_cvpr19}.
Top $k$ arguments from the heatmaps are selected to form the input tensor of $\mb{x}\in\mathbb{R}^{J \times 2k}$, and the $k$ arguments are randomly permuted during generation to prevent learning biases.
The input $\mb{x}$ is projected to an embedding $\mb{h}\in\mathbb{R}^{J \times m}$ and then processed by a graph neural network (GNN) that takes into account the inter-joint interactions via an learnable adjacency matrix $\mb{A}\in\mathbb{R}^{J \times J}$.
The condition $\mb{c}\in\mathbb{R}^d$ is computed as:
\begin{equation}
    \label{eq:gcn}
    \mb{c} = f_c(\textup{Flatten}(\sigma(\mb{A}\,f_{g}(\mb{h})))~,
\end{equation}
where $f_g: \mathbb{R}^{J \times m} \rightarrow \mathbb{R}^{J \times m}$ and $f_c: \mathbb{R}^{Jm} \rightarrow \mathbb{R}^{d}$ are learnable linear projections, $\sigma$ is the SiLU activation function, and the Flatten function maps $\mathbb{R}^{J \times m} \rightarrow \mathbb{R}^{Jm}$.
The condition $\mb{c}$ is concatenated with $\mb{q}$ and $t$ to construct the input tensor for the QDE function $f_{\textup{QDE}}$ in~\cref{eq:para_qde}.

\subsection{Shape estimation}
\label{subsec:method_shape}

The shape parameter $\beta\in\mathbb{R}^{10}$ controls the first 10 principle components (PC) of the SMPL mesh~\cite{loper_acm15}.
PC are linear variables, thus we apply the vanilla flow matching framework to the learning of $\beta$.
From the initial shape sampled from the Gaussian distribution with zero mean and identity variance, \ie $\beta_0 \sim \mathcal{N}(0,\mb{I})$, we learn the velocity flow $\dot{\beta}$, \ie shape shifting rate, that maps $\beta_0$ to the ground-truth ${\beta}_1$.
The estimation model is parameterized as the ODE: $\dot{\beta}_t = f_\textup{ODE}(\beta_t,t,\mb{c})$.
The OT of shape generation is the \texttt{lerp} between $\beta_0$ and ${\beta}_1$:
\begin{equation}
    \label{eq:lerp}
    \begin{split}
    & \beta_t = \textup{\texttt{lerp}}(\beta_0,\beta_1,t) = (1-t)\mb{\beta_0} + t\beta_1~, \\
    & \Rightarrow \dot{\beta}^* = \beta_1 - \beta_0~,
    \end{split}
\end{equation}
where $\dot{\beta}^*$ is the OT velocity.
The learning objective is the regression task to minimize the squared error:
\begin{equation}
     \mathcal{L_\beta} = \lVert f(\beta_t,t,\mb{c}) - \beta^* \rVert _2^2~, \, \forall t \in [0, 1]~.
\end{equation}

\subsection{Solving the differential equations}
\label{subsec:method_solvers}

Since the differential functions $f_{\textup{QDE}}$ and $f_{\textup{ODE}}$ are parameterized by neural networks, analytical solutions are not available, and numerical solvers are required.
The integration from \cref{eq:solution_qde} is also referred as the Euler method on a quaternion-valued ODE, \ie the first-order solver following the Runge-Kutta scheme (RK).
Higher-order solvers do not necessarily result in more accurate solutions, yet significantly increase the computation~\cite{le_tmlr26}.
The best trade-off is often recorded with the 2\textsuperscript{nd}-order solver~\cite{le_iclr26,vuik_numericalode23}, and we modify the integration scheme to handle the QDE as:
\begin{equation}
    \label{eq:rk2_qde}
    \begin{split}
        & \mb{q}_{t+\frac{\Delta t}{2}} = \mb{q}_t \otimes \textup{quat}_{\frac{\Delta t}{2}}(f_{\textup{QDE}}(\mb{q}_t,t,\mb{c})), \\
        & \mb{q}_{t+\Delta t} = \mb{q}_{t+\frac{\Delta t}{2}} \otimes \textup{quat}_{\Delta t}(f_{\textup{QDE}}(\mb{q}_{t+\frac{\Delta t}{2}},t+\frac{\Delta t}{2},\mb{c})).
    \end{split}
\end{equation}
To solve the ODE of the shape parameter $\beta$, we apply the vanilla RK2 solver~\cite{chen_nips18}:
\begin{equation}
    \label{eq:rk2_ode}
    \beta_{t+\Delta t} = \beta_t + f_{\textup{ODE}}\left(\beta_t + f_{\textup{ODE}}(\beta_t,t,\mb{c})\frac{\Delta t}{2}, t+\frac{\Delta t}{2}, \mb{c}\right)\Delta t.
\end{equation}
The predicted pose $\mb{q}_1$ and shape $\beta_1$ are iteratively solved from the initial $\mb{q}_0$ and $\beta_0$ using \cref{eq:rk2_qde,eq:rk2_ode}, which, as we show in Section~\ref{sec:experiments}, result in smooth CNFs following OT trajectories.

\subsection{Learning objectives}
\label{subsec:method_objectives}

To utilize the human kinematics constraint of the SMPL model, we obtain the predicted pose $\hat{\mb{q}}_1$ and shape $\hat{\beta}_1$ by linear interpolation with a constant-velocity assumption:
\begin{equation}
    \label{eq:predictions}
    \begin{split}
        & \hat{\mb{q}}_1 = \mb{q}_t \otimes \textstyle\left(\cos{\frac{\lVert\omega_t\rVert(1-t)}{2},\;\sin{\frac{\lVert\omega_t\rVert(1-t)}{2}} \; \frac{{\omega}_{t}}{\lVert \omega_{t} \rVert}}\right), \\
        & \hat{\beta}_1 = \beta_t + (1-t)\beta'_t~,
        \quad \textup{with}~t \sim \mathcal{U}(0,1)~.
    \end{split}
\end{equation}

We apply the predictions to the SMPL model via the forward function~\cite{loper_acm15}, obtaining the predicted mesh $V\in\mathbb{R}^{6980\times3}$.
A linear joint regressor~\cite{choi_eccv20,loper_acm15} is used to project the $6980$ vertices to the 3D skeleton with $J$ body joints, \ie $\hat{V}\in\mathbb{R}^{6980\times3} \rightarrow [\hat{y_j}]_{j=1}^J\in\mathbb{R}^{J\times3}$.
The keypoint loss is the L1 distance between the prediction  $\hat{y} \in \mathbb{R}^3$ and the ground-truth 3D keypoints $y$, \ie $\mathcal{L}_{\textup{y}} = |\hat{y} - y|$.
The total loss is the aggregation of all the objectives over all $J$ joints:
\begin{equation}
    \label{eq:loss}
    \mathcal{L} = \lambda_\textup{y}\sum_{j=1}^{J}\mathcal{L}_{\textup{y}j}+ \lambda_\omega\sum_{j=1}^{J}\mathcal{L}_{\omega j} + \lambda_\beta\mathcal{L}_{\beta}~,
\end{equation}
where $\lambda_\textup{y},\lambda_{\omega},\lambda_{\beta}$ are hyperparameters that scale the contributions of the respective losses.

\subsection{Implementation details}
\label{subsec:method_implemetation}

CQF-HMR is designed to be an end-to-end single-frame 3D human mesh recovery approach.
The DENet, which predicts the angular velocity $\omega$ and shape flow $\dot{\beta}$, is a two-layer MLP with a hidden dimension of 1024, followed by LayerNorms and SiLU activations~\cite{elfwing_nn18}.
To obtain pseudo-ground-truth SMPL parameters for the realization of the OT quaternion flows~\cref{eq:slerp}, we fit the SMPL model to the ground-truth 3D keypoints via an optimization process~\cite{pavlakos_cvpr19}.
From the extracted 2D joint heatmaps from HRNet~\cite{sun_cvpr19}, we collect the top-$k$ candidates by ranking their confidence scores, then normalize the top-$k$ 2D poses to root-origin and identity standard deviation.
We learn the 2D lifting condition $\mb{c}$ via a GCN module, with the skeleton adjacency initialized to zeros~\cite{le_tmlr26}, using the top-$k$ poses as input.

We train the CQF-HMR on the Human3.6M dataset for $100$ epochs, with a batch size of $256$, the AdamW optimizer, and a learning rate of $3\times10^{-4}$ scheduled to reduce by a factor $0.1$ at epoch $90$.
The loss weighting coefficients are chosen as $\lambda_\textup{y}=\lambda_{\omega}=1,\lambda_{\beta}=1\times10^{-2}$.
We finetune CQF-HMR on the 3DPW using the same hyperparameters with the one-cycle learning rate scheduler.
At inference, we implement the RK2 solver for solving both the continuous quaternion and shape normalizing flows.
The flows are bounded within the range of $t\in[0,1]$ with a total of $20$ time steps, and the step size is correspondingly $\Delta t = \frac{1}{20}$.
All experiments are conducted on three random seeds $0,1,2$, and we report the mean and standard deviations.
\section{Experiments}
\label{sec:experiments}

\subsection{Datasets}
\label{subsec:exp_datasets}

We evaluate CQF-HMR on three established datasets.
The first dataset is the Human3.6M~\cite{ionescu_pami14} containing $7$ actors performing actions captured from $4$ different camera views.
Following prior work~\cite{wehrbein_iccv21,holmquist_iccv23,le_iclr26}, we train on subjects S1, S5, S6, S7, S8 and evaluate on S9 and S11, across all camera views.
For comparison with related methods, we consider two evaluation setups: 1) on every $64^{\textup{th}}$ of the regular test set; and 2) the highly ambiguous subset defined by~\citet{wehrbein_iccv21}, where a pose is ambiguous if the fitted 2D Gaussian on the heatmaps has a width greater than $5$ pixels.
The second more challenging dataset is the 3DPW which contains complex outdoor scenes and heavy occlusions.
We finetune the pre-trained CQF-HMR on the training split of 3DPW and evaluate on the test split.
The last dataset is the EMDB~\cite{kaufmann_iccv23}, specifically the subset defined by~\citet{wehrbein_wacv25}, which contains human poses captured from a moving camera with occlusion.

\subsection{Evaluation metrics}
\label{subsec:exp_metrics}

We follow the standard 3D HPE evaluation protocols.
The first protocol measures the mean Euclidean distance between the predicted 3D keypoints to the ground-truths, namely Mean Per Joint Position Error (MPJPE).
The second protocol adopts MPJPE calculation with Procrustes Alignment procedure (PA-MPJPE) to eliminate the global translation, rotation, and scale differences, only focusing on the relative shape and pose discrepancy.
The two MPJPE metrics are measured with 3D keypoints in \unit{mm}.
On the ambiguous set of Human3.6M, we also report: 1) the Percentage of Correct Keypoints (PCK) that measures the percentage (in $\%$) of predicted keypoints that are within a distance of 150mm with respect to the corresponding ground-truths; and 2) the Correct Poses Score (CPS) that measures the area under the curve for a range of threshold from $0$\unit{mm} to $300$\unit{mm}.
The CPS only counts a pose to be correct if all joint errors are lower than the threshold.
On 3DPW, in addition to MPJPE and PA-MPJPE, we also compute the Per Vertex Error (PVE) between the predicted SMPL meshes and ground truths.
As a probabilistic approach, we quantify the multi-hypothesis generation with the Diversity metric that computes the average distance from all joint hypotheses to their mean position, for visible and invisible joints.

\subsection{Comparisons to related work}
\label{subsec:exp_comparisons}

\begin{table}[t]
    \centering
    \setlength\tabcolsep{1pt}
    \begin{tabularx}{\linewidth}{@{}lXccccc}
    \toprule
    Method & & \footnotesize{Single} & \footnotesize{Mesh} & \footnotesize{Prob.} & \footnotesize{MPJPE~$\downarrow$} & \footnotesize{PA-MPJPE~$\downarrow$} \\
    \midrule
    PoseFormerV2~\cite{zhao_cvpr23} & & \xmark & \xmark & \xmark & $45.2$ & $35.6$ \\
    MHFormer~\cite{li_cvpr2022} & & \xmark & \xmark & \cmark & $43.0$ & $34.6$ \\
    ManiPose~\cite{rommel_nips24} & & \xmark & \xmark & \cmark & $39.1$ & $34.1$ \\
    VIBE~\cite{kocabas_cvpr20} & & \xmark & \cmark & \xmark & $65.9$ & $41.5$ \\
    SPIN~\cite{kolotouros_iccv19} & & \cmark & \cmark & \xmark & $62.5$ & $41.1$ \\
    HybrIK~\cite{li_hybrik21} & & \cmark & \cmark & \xmark & $55.4$ & $33.6$ \\
    HMR2.0~\cite{goel_iccv23} & & \cmark & \cmark & \xmark & $44.8$ & $33.6$ \\
    \citet{wehrbein_iccv21} & & \cmark & \xmark & \xmark & $61.8$ & $43.8$ \\
    DiffPose~\cite{holmquist_iccv23} & & \cmark & \xmark & \xmark & $64.5$ & $45.2$ \\
    FMPose~\cite{le_tmlr26} & & \cmark & \xmark & \xmark & $58.9{\scriptstyle\mathrm{\pm0.4}}$ & $41.5{\scriptstyle\mathrm{\pm0.1}}$ \\
    \rowcolor{gray!10}
    CQF-HMR~\textsubscript{($\mathcal{N}$)} & & \cmark & \cmark & \xmark & $59.4{\scriptstyle\mathrm{\pm0.2}}$ & $39.5{\scriptstyle\mathrm{\pm0.3}}$ \\
    \rowcolor{gray!10}
    CQF-HMR~\textsubscript{($\mathcal{D}$)} & & \cmark & \cmark & \xmark & $59.0{\scriptstyle\mathrm{\pm0.3}}$ & $40.3{\scriptstyle\mathrm{\pm0.6}}$ \\
    FMPose3D~\cite{wang_cvpr26} & & \cmark & \xmark & \cmark & $47.3$ & $38.3$ \\
    GraphMDN~\cite{oikarinen_gmdn21} & & \cmark & \xmark & \cmark & $46.2$ & $36.3$ \\
    \citet{wehrbein_iccv21} & & \cmark & \xmark & \cmark & $44.3$ & $32.4$ \\
    DiffPose~\cite{holmquist_iccv23} & & \cmark & \xmark & \cmark & $44.2{\scriptstyle\mathrm{\pm0.2}}$ & $32.1{\scriptstyle\mathrm{\pm0.1}}$ \\
    FMPose~\cite{le_tmlr26} & & \cmark & \xmark & \cmark & $41.7{\scriptstyle\mathrm{\pm0.3}}$ & $30.6{\scriptstyle\mathrm{\pm0.1}}$ \\
    \rowcolor{gray!10}
    CQF-HMR~\textsubscript{($\mathcal{N}$)} & & \cmark & \cmark & \cmark & $41.1{\scriptstyle\mathrm{\pm0.3}}$ & $28.5{\scriptstyle\mathrm{\pm0.2}}$ \\
    \rowcolor{gray!10}
    CQF-HMR~\textsubscript{($\mathcal{D}$)} & & \cmark & \cmark & \cmark & $39.4{\scriptstyle\mathrm{\pm0.2}}$ & $27.9{\scriptstyle\mathrm{\pm0.1}}$ \\
    \bottomrule
    \end{tabularx}
    \caption{
    Quantitative comparisons to related work on the regular Human3.6M testing split.
    Single: methods that use a single input image.
    Mesh: SMPL-based method.
    Prob.: Probabilistic approach, \ie multi-hypothesis generation.
    \colorbox{gray!10}{Our method}.
    The subscripts \textsubscript{($\mathcal{N}$)} and \textsubscript{($\mathcal{D}$)} denote the sampling of initial pose from the Gaussian distribution or the human prior VPoser~\cite{pavlakos_cvpr19} respectively.
    }
    \label{tab:exp_h36m}
\end{table}

\textbf{On Human3.6M}, we first evaluate CQF-HMR on the action sequences of S9 and S11.
Together with related work, the results are collected in~\cref{tab:exp_h36m}.
We group the methods based on three categories: 1) Using single-image input (Single), 2) SMPL-based mesh output (Mesh), and 3) Probabilistic approach that generates multiple hypotheses (Prob.).
All Mesh methods predict the SMPL's parameters and require the Human3.6M joint regressor to obtain the standard $17$ joints for evaluation.
The results of probabilistic approaches are recorded as the minimum error of $200$ hypotheses.
To make probabilistic methods deterministic, we sample the initial pose at zeros, \ie origin for 3D-keypoint approaches~\cite{wehrbein_iccv21,holmquist_iccv23,le_tmlr26} or neutral SMPL pose (CQF-HMR).

Overall, our CQF-HMR shows clear advantages over related work.
Compared to the closest probabilistic baseline FMPose~\cite{le_tmlr26}, CQF-HMR~\textsubscript{($\mathcal{N}$)} improves the MPJPE by $1.4\%~(0.6\unit{mm})$ and the PA-MPJPE by $6.8\%~(2.1\unit{mm})$ respectively, whereas CQF-HMR~\textsubscript{($\mathcal{N}$)} improves the MPJPE by $5.7\%~(2.4\unit{mm})$ and the PA-MPJPE by $8.8\%~(2.7\unit{mm})$.
The performance boost of sampling from the human prior VPoser is resulted from the more plausible initial human poses, which significantly reduces the learning and inference of OT distribution mappings, compared to random and highly implausible initial poses sampled from the Gaussian distribution.
Regarding the deterministic setting, both CQF-HMR models achieve on-par MPJPE as FMPose, but outperform on the PA-MPJPE with CQF-HMR~\textsubscript{($\mathcal{N}$)} decreases by $4.8\%~(2.0\unit{mm})$ and CQF-HMR~\textsubscript{($\mathcal{D}$)} by $2.8\%~(1.2\unit{mm})$.

\begin{table}[t]
    \centering
    \setlength\tabcolsep{0.5pt}
    \begin{tabularx}{\linewidth}{@{}lXcccc}
    \toprule
    Method & & \footnotesize{MPJPE~$\downarrow$} & \footnotesize{PA-MPJPE~$\downarrow$} & \footnotesize{PCK~$\uparrow$} & \footnotesize{CPS~$\uparrow$} \\
    \midrule
    \citet{li_cvpr19} & & $81.1$ & $66.0$ & $85.7$ & $119.9$ \\
    \citet{sharma_iccv19} & & $78.3$ & $61.1$ & $88.5$ & $136.4$ \\
    \citet{wehrbein_iccv21} & & $71.0$ & $54.2$ & $93.4$ & $171.0$ \\
    DiffPose~\cite{holmquist_iccv23} & & $66.5{\scriptstyle\mathrm{\pm1.4}}$ & $48.5{\scriptstyle\mathrm{\pm0.2}}$ & $94.3{\scriptstyle\mathrm{\pm0.1}}$ & $194.2{\scriptstyle\mathrm{\pm2.2}}$ \\
    FMPose~\cite{le_tmlr26} & & $60.0{\scriptstyle\mathrm{\pm0.3}}$ & $47.5{\scriptstyle\mathrm{\pm0.3}}$ & $96.0{\scriptstyle\mathrm{\pm0.1}}$ & $187.6{\scriptstyle\mathrm{\pm1.6}}$ \\
    FMPose\textsuperscript{\dag}~\cite{le_tmlr26} & & $58.9{\scriptstyle\mathrm{\pm0.1}}$ & $46.7{\scriptstyle\mathrm{\pm0.2}}$ & $96.6{\scriptstyle\mathrm{\pm0.1}}$ & $194.7{\scriptstyle\mathrm{\pm0.9}}$ \\
    \rowcolor{gray!10}
    CQF-HMR~\textsubscript{($\mathcal{N}$)} & & $56.3{\scriptstyle\mathrm{\pm0.4}}$ & $42.5{\scriptstyle\mathrm{\pm0.2}}$ & $96.4{\scriptstyle\mathrm{\pm0.1}}$ & $210.5{\scriptstyle\mathrm{\pm0.5}}$ \\
    \rowcolor{gray!10}
    CQF-HMR~\textsubscript{($\mathcal{D}$)} & & $54.4{\scriptstyle\mathrm{\pm0.3}}$ & $42.5{\scriptstyle\mathrm{\pm0.2}}$ & $96.2{\scriptstyle\mathrm{\pm0.1}}$ & $205.6{\scriptstyle\mathrm{\pm0.7}}$ \\
    \bottomrule
    \end{tabularx}
    \caption{
    Quantitative comparisons on the Human3.6M ambiguous split defined by~\citet{wehrbein_iccv21}.
    \textsuperscript{\dag} denotes the FMPose~\cite{le_tmlr26} version that uses random sampling strategy from DiffPose~\cite{holmquist_iccv23}.
    }
    \label{tab:exp_ah36m}
\end{table}

Following~\cite{wehrbein_iccv21,holmquist_iccv23,le_tmlr26}, we also evaluate CQF-HMR on the ambiguous subset that contains only challenging occluded poses, and the results are collected in~\cref{tab:exp_ah36m}.
Compared to the alternative version of FMPose~\cite{le_tmlr26} that uses the random heatmap sampling strategy from DiffPose~\cite{holmquist_iccv23}, CQF-HMR~\textsubscript{($\mathcal{N}$)} improves the MPJPE by $4.4\%~(2.6\unit{mm})$, and the PA-MPJPE by $8.9\%~(4.2\unit{mm})$; while CQF-HMR~\textsubscript{($\mathcal{D}$)} further increase the performance with $7.6\%~(4.5\unit{mm})$ decrease in MPJPE.
While having similar PCK with sub $1\%$ differences to~\cite{holmquist_iccv23,le_tmlr26}, CQF-HMR achieving higher CPS demonstrates the better coverage of CQF-HMR's predictions on the ambiguous poses.

\begin{table}[t]
    \centering
    \setlength\tabcolsep{2pt}
    \begin{tabularx}{\linewidth}{@{}lcXccc}
    \toprule
    Method & \footnotesize{Prob.} & & \footnotesize{MPJPE~$\downarrow$} & \footnotesize{PA-MPJPE~$\downarrow$} & \footnotesize{PVE~$\downarrow$} \\
    \midrule
    SPIN~\cite{kolotouros_iccv19} & \xmark & & $96.9$ & $59.0$ & $116.4$ \\
    HybrIk~\cite{li_hybrik21} & \xmark & & $74.1$ & $45.0$ & $94.5$ \\
    CLIFF~\cite{li_cliff22} & \xmark & & $69.0$ & $43.0$ & $81.2$ \\
    HMR2.0~\cite{goel_iccv23} & \xmark & & $70.0$ & $44.5$ & \xmark \\
    ProHMR~\cite{kolotouros_iccv21} & \cmark & & $81.5$ & $48.2$ & \xmark \\
    HierProbHuman~\cite{sengupta_iccv21} & \cmark & & $70.9$ & $43.8$ & \xmark \\
    HuManiFlow~\cite{sengupta_cvpr23} & \cmark & & $65.1$ & $39.9$ & $75.5$ \\
    ScoreHypo~\cite{xu_cvpr24} & \cmark & & $63.0$ & $ 37.6$ & $73.4$ \\
    \citet{wehrbein_wacv25}\textsuperscript{\ddag} & \cmark & & $46.2$ & $29.8$ & $54.4$ \\
    \rowcolor{gray!10}
    CQF-HMR~\textsubscript{($\mathcal{N}$)} & \cmark & & $61.5{\scriptstyle\mathrm{\pm0.3}}$ & $38.2{\scriptstyle\mathrm{\pm0.2}}$ & $78.8{\scriptstyle\mathrm{\pm0.4}}$ \\
    \rowcolor{gray!10}
    CQF-HMR~\textsubscript{($\mathcal{D}$)} & \cmark & & $64.1{\scriptstyle\mathrm{\pm0.2}}$ & $40.9{\scriptstyle\mathrm{\pm0.1}}$ & $82.7{\scriptstyle\mathrm{\pm0.4}}$ \\
    \bottomrule
    \end{tabularx}
    \caption{
    Quantitative results on the 3DPW dataset.
    \textsuperscript{\ddag} The method from~\cite{wehrbein_wacv25} was trained on different database, BEDLAM~\cite{black_bedlam_cvpr23} + AGORA~\cite{patel_agora_cvpr21}, which are significantly larger than the standard training protocol of Human3.6M~\cite{ionescu_pami14}.
    }
    \label{tab:exp_3dpw}
\end{table}

\begin{table}[t]
    \centering
    \setlength\tabcolsep{2pt}
    \begin{tabularx}{\linewidth}{@{}lXccc}
    \toprule
    Method & & \footnotesize{MPJPE~$\downarrow$} & \footnotesize{PA-MPJPE~$\downarrow$} & \footnotesize{PVE~$\downarrow$} \\
    \midrule
    HuManiFlow~\cite{sengupta_cvpr23} & & $88.7$ & $56.5$ & $100.6$ \\
    ScoreHypo~\cite{xu_cvpr24} & & $87.4$ & $58.5$ & $99.6$ \\
    \citet{wehrbein_wacv25}\textsuperscript{\ddag} & & $63.6$ & $40.9$ & $72.0$ \\
    \rowcolor{gray!10}
    CQF-HMR~\textsubscript{($\mathcal{N}$)} & & $88.3{\scriptstyle\mathrm{\pm0.9}}$ & $56.5{\scriptstyle\mathrm{\pm0.2}}$ & $112.1{\scriptstyle\mathrm{\pm0.8}}$ \\
    \rowcolor{gray!10}
    CQF-HMR~\textsubscript{($\mathcal{D}$)} & & $92.4{\scriptstyle\mathrm{\pm0.2}}$ & $60.9{\scriptstyle\mathrm{\pm0.2}}$ & $119.0{\scriptstyle\mathrm{\pm0.6}}$ \\
    \bottomrule
    \end{tabularx}
    \caption{
    Quantitative results on the EMDB dataset.
    \textsuperscript{$\ddag$} \cite{wehrbein_iccv21} does not follow the same training protocol.
    }
    \label{tab:exp_emdb}
\end{table}

\begin{figure*}[t]
    \centering
    \begin{subfigure}[b]{0.46\textwidth}
        \centering
        \includegraphics[trim= 0 10 0 20, clip, width=\linewidth]{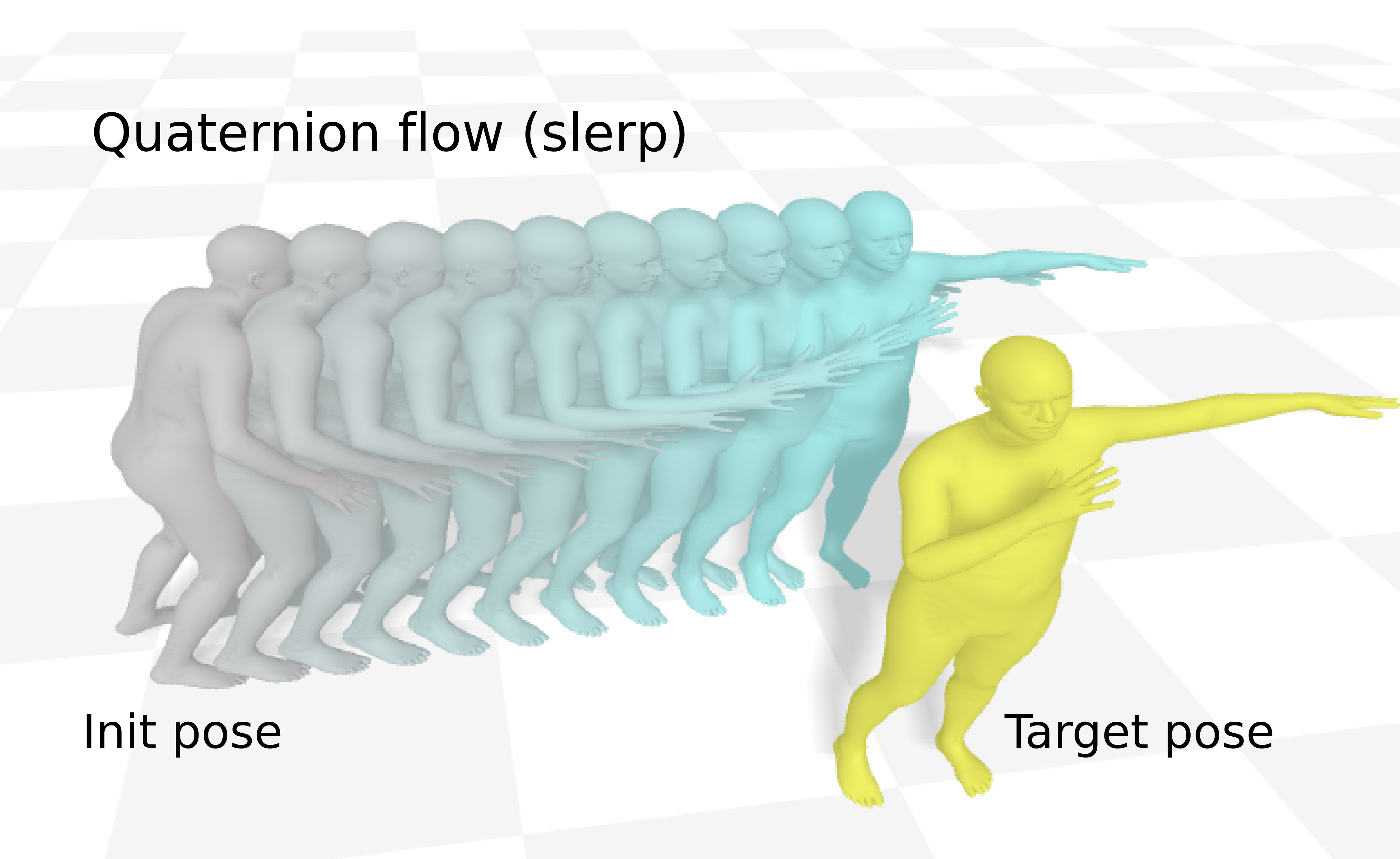}
    \end{subfigure}%
    ~ 
    \begin{subfigure}[b]{0.46\textwidth}
        \centering
        \includegraphics[trim= 0 10 0 20, clip, width=\linewidth]{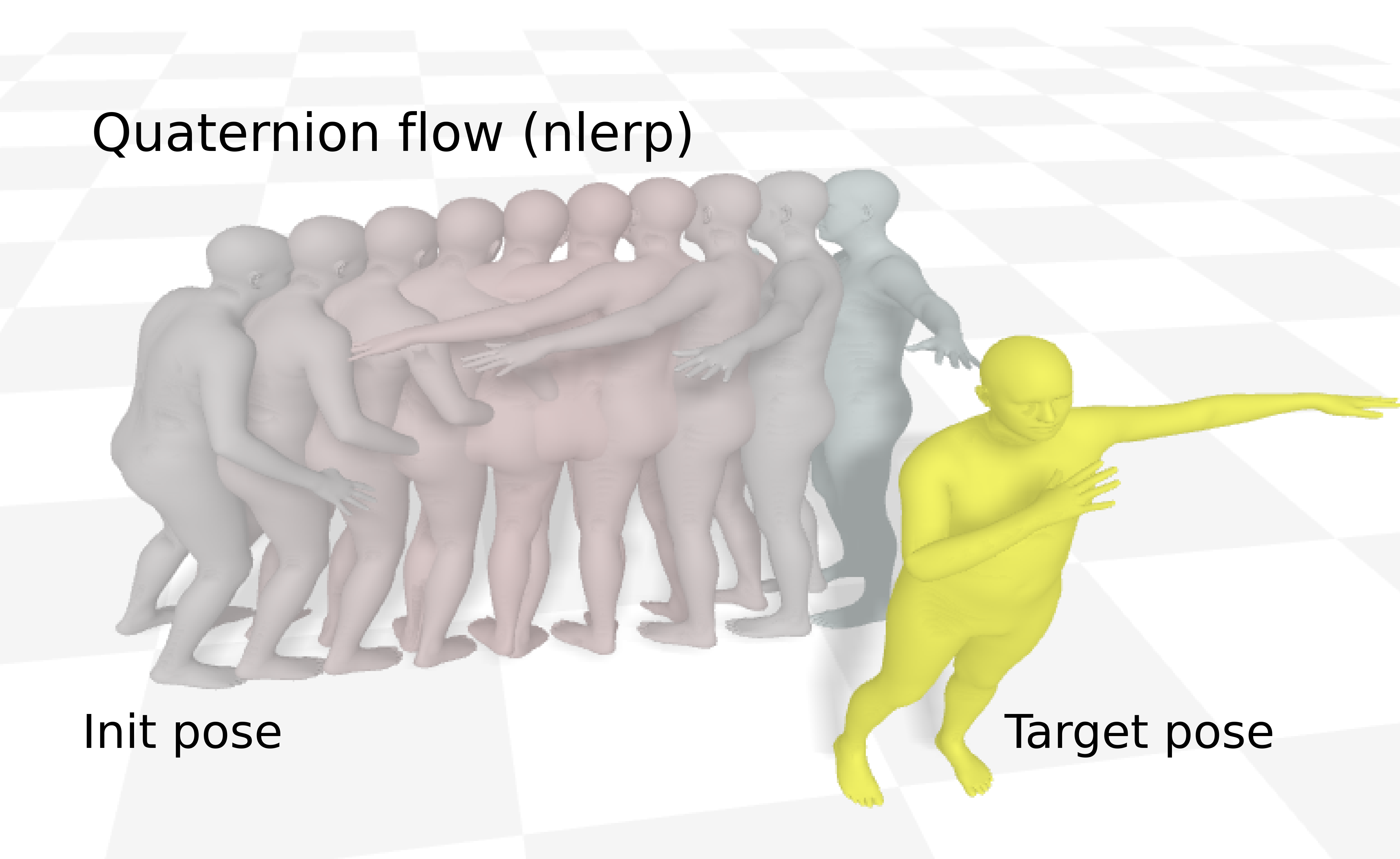}
    \end{subfigure}
    \\
    \begin{subfigure}[b]{0.46\textwidth}
        \centering
        \includegraphics[trim= 0 10 0 20, clip, width=\linewidth]{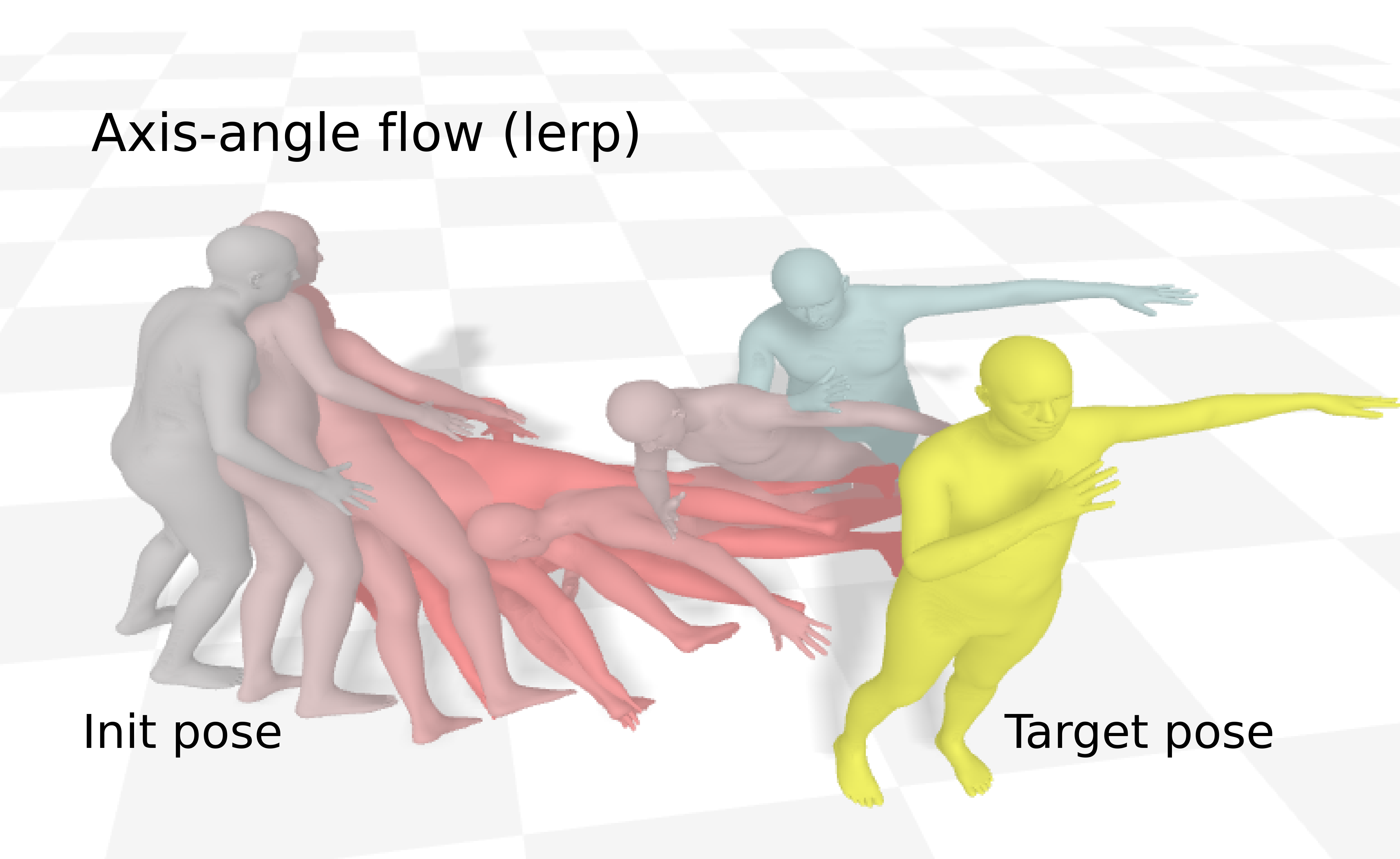}
    \end{subfigure}%
    ~ 
    \begin{subfigure}[b]{0.46\textwidth}
        \centering
        \includegraphics[trim= 0 10 0 20, clip, width=\linewidth]{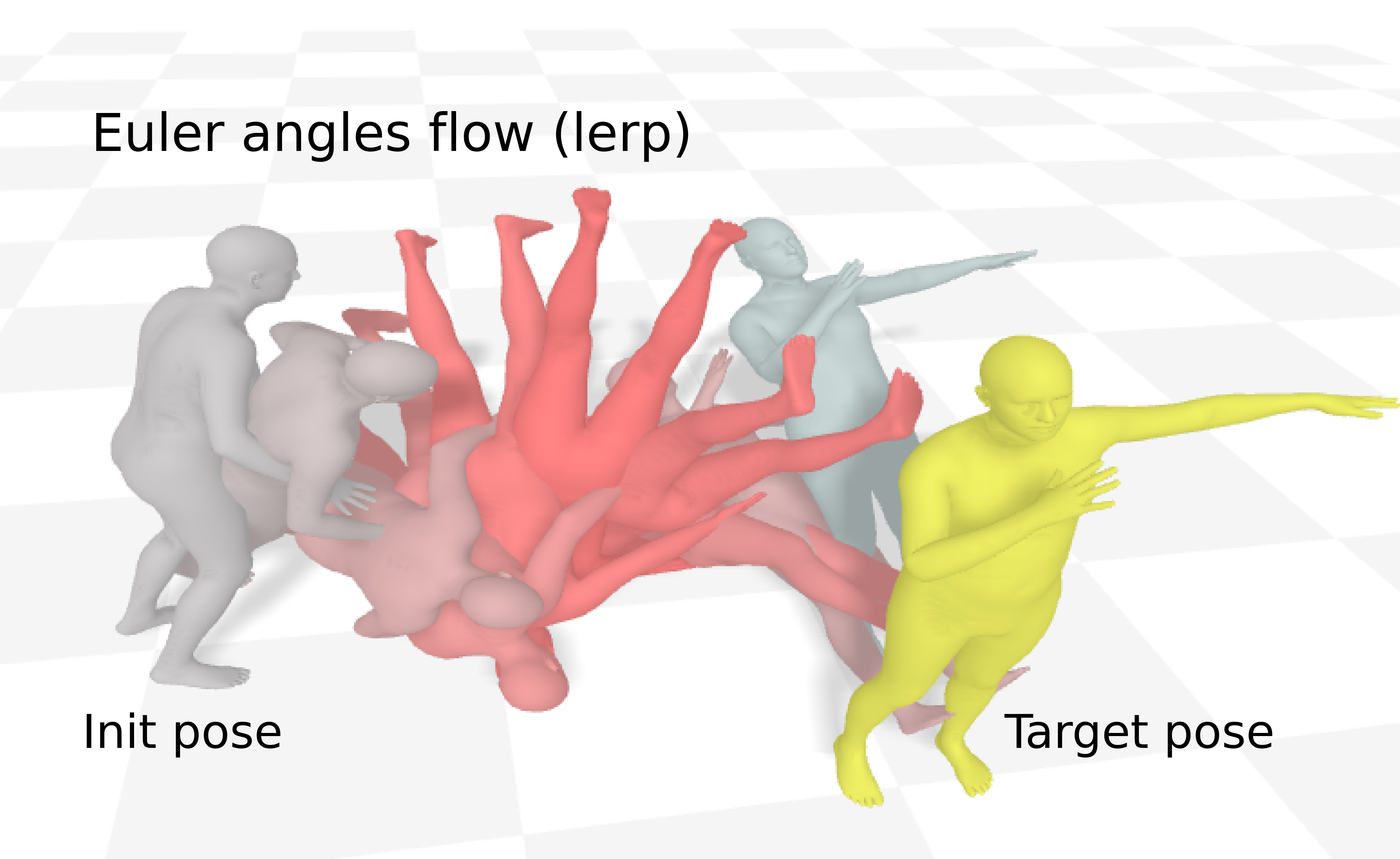}
    \end{subfigure}
    \caption{
    The comparison between the OT flows of the 3D rotations: quaternion with \texttt{slerp}, quaternion with \texttt{nlerp} (normalized \texttt{lerp}), Euler angles with \texttt{lerp}, and Axis-angle with \texttt{lerp}.
    The yellow mesh is the ground-truth pose from the Human3.6M test split.
    \textcolor{Cyan}{Blue} indicates low MPJPE, while \textcolor{Red}{red} indicates high MPJPE.
    The init pose is sampled from the VPoser distribution~\cite{pavlakos_cvpr19}, and the normalizing flow direction is from left to right.
    The trajectories are obtained by solving the ODEs with their respective OT derivatives.
    }
    \label{fig:rotations}
\end{figure*}

\begin{table*}[t]
    \centering
    \small
    \setlength\tabcolsep{1pt}
    \begin{tabularx}{\linewidth}{@{\extracolsep{\fill}}lcccccccccc}
    \toprule
    \multirow{3}{*}{Rotation} & \multirow{3}{*}{Initial} & \multirow{3}{*}{OT flow} & 
    \multicolumn{4}{c}{Deterministic (best of $1$)} & \multicolumn{4}{c}{Probabilistic (best of $100$)} \\
    & & & \multicolumn{2}{c}{H36M} & \multicolumn{2}{c}{Ambiguous H36M} & \multicolumn{2}{c}{H36M} & \multicolumn{2}{c}{Ambiguous H36M} \\
    & & & MPJPE~$\downarrow$ & PA-MPJPE~$\downarrow$ & MPJPE~$\downarrow$ & PA-MPJPE~$\downarrow$ & MPJPE~$\downarrow$ & PA-MPJPE~$\downarrow$ & MPJPE~$\downarrow$ & PA-MPJPE~$\downarrow$ \\
    \midrule
    Axis-angle & $\mathcal{N}$      & \texttt{lerp} & $113.4{\scriptstyle\mathrm{\pm9.1}}$ & $43.3{\scriptstyle\mathrm{\pm0.5}}$ & $151.3{\scriptstyle\mathrm{\pm10.0}}$ & $64.9{\scriptstyle\mathrm{\pm0.3}}$ & $45.4{\scriptstyle\mathrm{\pm0.3}}$ & $30.6{\scriptstyle\mathrm{\pm0.1}}$ & $61.6{\scriptstyle\mathrm{\pm0.1}}$ & $45.1{\scriptstyle\mathrm{\pm0.3}}$ \\
    Euler angles & $\mathcal{N}$    & \texttt{lerp} & $64.9{\scriptstyle\mathrm{\pm0.5}}$ & $40.9{\scriptstyle\mathrm{\pm0.1}}$ & $90.6{\scriptstyle\mathrm{\pm2.3}}$ & $61.1{\scriptstyle\mathrm{\pm0.3}}$ & $43.9{\scriptstyle\mathrm{\pm0.2}}$ & $30.1{\scriptstyle\mathrm{\pm0.1}}$ & $60.1{\scriptstyle\mathrm{\pm0.1}}$ & $44.5{\scriptstyle\mathrm{\pm0.2}}$ \\
    Quaternion & $\mathcal{N}$      & \texttt{lerp} & $74.7{\scriptstyle\mathrm{\pm4.4}}$ & $41.6{\scriptstyle\mathrm{\pm0.1}}$ & $91.8{\scriptstyle\mathrm{\pm1.9}}$ & $61.8{\scriptstyle\mathrm{\pm0.8}}$ & $45.7{\scriptstyle\mathrm{\pm0.4}}$ & $31.9{\scriptstyle\mathrm{\pm0.2}}$ & $61.5{\scriptstyle\mathrm{\pm0.2}}$ & $46.5{\scriptstyle\mathrm{\pm0.4}}$ \\
    Quaternion & $\mathcal{N}$      & \texttt{nlerp} & $121.3{\scriptstyle\mathrm{\pm5.1}}$ & $47.3{\scriptstyle\mathrm{\pm0.4}}$ & $116.0{\scriptstyle\mathrm{\pm4.5}}$ & $68.1{\scriptstyle\mathrm{\pm1.7}}$ & $68.0{\scriptstyle\mathrm{\pm0.6}}$ & $45.1{\scriptstyle\mathrm{\pm0.4}}$ & $81.7{\scriptstyle\mathrm{\pm0.5}}$ & ${58.9\scriptstyle\mathrm{\pm0.6}}$ \\
    Quaternion & $\mathcal{N}$      & \texttt{slerp} & $59.4{\scriptstyle\mathrm{\pm0.2}}$ & $39.6{\scriptstyle\mathrm{\pm0.2}}$ & $77.4{\scriptstyle\mathrm{\pm0.8}}$ & $59.4{\scriptstyle\mathrm{\pm0.4}}$ & $43.5{\scriptstyle\mathrm{\pm0.3}}$ & $30.2{\scriptstyle\mathrm{\pm0.1}}$ & $59.3{\scriptstyle\mathrm{\pm0.3}}$ & $45.0{\scriptstyle\mathrm{\pm0.1}}$ \\
    \midrule
    Axis-angle & $\mathcal{D}$      & \texttt{lerp} & $89.3{\scriptstyle\mathrm{\pm2.6}}$ & $46.1{\scriptstyle\mathrm{\pm0.2}}$ & $116.6{\scriptstyle\mathrm{\pm6.6}}$ & $67.0{\scriptstyle\mathrm{\pm0.5}}$ & $48.1{\scriptstyle\mathrm{\pm0.3}}$ & $33.5{\scriptstyle\mathrm{\pm0.1}}$ & $64.9{\scriptstyle\mathrm{\pm0.2}}$ & $49.6{\scriptstyle\mathrm{\pm0.2}}$ \\
    Euler angles & $\mathcal{D}$    & \texttt{lerp} & $64.5{\scriptstyle\mathrm{\pm0.3}}$ & $42.3{\scriptstyle\mathrm{\pm0.2}}$ & $92.1{\scriptstyle\mathrm{\pm3.3}}$ & $64.4{\scriptstyle\mathrm{\pm1.7}}$ & $42.6{\scriptstyle\mathrm{\pm0.2}}$ & $30.4{\scriptstyle\mathrm{\pm0.1}}$ & $59.9{\scriptstyle\mathrm{\pm0.3}}$ & $45.9{\scriptstyle\mathrm{\pm0.1}}$ \\
    Quaternion & $\mathcal{D}$      & \texttt{lerp} & $72.4{\scriptstyle\mathrm{\pm1.9}}$ & $42.5{\scriptstyle\mathrm{\pm0.3}}$ & $92.7{\scriptstyle\mathrm{\pm3.2}}$ & $61.5{\scriptstyle\mathrm{\pm0.8}}$ & $46.1{\scriptstyle\mathrm{\pm0.3}}$ & $32.7{\scriptstyle\mathrm{\pm0.1}}$ & $62.0{\scriptstyle\mathrm{\pm0.2}}$ & $47.8{\scriptstyle\mathrm{\pm0.2}}$ \\
    Quaternion & $\mathcal{D}$      & \texttt{nlerp} & $125.7{\scriptstyle\mathrm{\pm9.1}}$ & $44.0{\scriptstyle\mathrm{\pm0.3}}$ & $111.7{\scriptstyle\mathrm{\pm6.8}}$ & $62.1{\scriptstyle\mathrm{\pm0.2}}$ & $48.9{\scriptstyle\mathrm{\pm0.3}}$ & $33.1{\scriptstyle\mathrm{\pm0.2}}$ & $63.5{\scriptstyle\mathrm{\pm0.2}}$ & $48.3{\scriptstyle\mathrm{\pm0.1}}$ \\
    Quaternion & $\mathcal{D}$      & \texttt{slerp} & $59.0{\scriptstyle\mathrm{\pm0.3}}$ & $40.3{\scriptstyle\mathrm{\pm0.6}}$ & $75.4{\scriptstyle\mathrm{\pm0.5}}$ & $58.3{\scriptstyle\mathrm{\pm0.7}}$ & $41.3{\scriptstyle\mathrm{\pm0.2}}$ & $29.3{\scriptstyle\mathrm{\pm0.2}}$ & $56.7{\scriptstyle\mathrm{\pm0.3}}$ & $44.3{\scriptstyle\mathrm{\pm0.3}}$ \\
    \bottomrule
    \end{tabularx}
    \caption{
    Ablation study on different body joint 3D rotations.
    The models are trained with their respective OT flows, \eg \texttt{lerp}, \texttt{nlerp}, \texttt{slerp}.
    Initial $\mathcal{N}$ means the initial poses are sampled from the Gaussian distribution, while $\mathcal{D}$ means sampling from VPoser~\cite{pavlakos_cvpr19}.
    }
    \label{tab:abl_rot3D}
\end{table*}

\begin{table}
    \centering
    \begin{tabular}{lcc}
        \toprule
        \multirow{2}{*}{Method} & \multicolumn{2}{c}{Diversity} \\
        & Visible & Invisible \\
        \midrule
        ProHMR~\cite{kolotouros_iccv21} & $35.1$ & $60.8$ \\
        HierProbHuman~\cite{sengupta_iccv21} & $47.6$ & $101.4$ \\
        HuManiFlow~\cite{sengupta_cvpr23} & $42.8$ & $116.0$ \\
        \citet{wehrbein_wacv25} & $35.3$ & $80.0$ \\
        \rowcolor{gray!10}
        CQF-HMR~\textsubscript{($\mathcal{N}$)} & $55.2{\scriptstyle\mathrm{\pm0.1}}$ & $89.5{\scriptstyle\mathrm{\pm0.5}}$ \\
        \rowcolor{gray!10}
        CQF-HMR~\textsubscript{($\mathcal{D}$)} & $40.2{\scriptstyle\mathrm{\pm0.5}}$ & $62.2{\scriptstyle\mathrm{\pm0.2}}$ \\
        \bottomrule
    \end{tabular}
    \caption{
    The Diversity measurements on the 3DPW dataset.
    }
    \label{tab:3dpw_div}
\end{table}

\begin{figure*}[t]
    \centering
    \begin{subfigure}[t]{0.24\textwidth}
        \centering
        \includegraphics[trim= 20 20 20 20, clip, width=\linewidth]{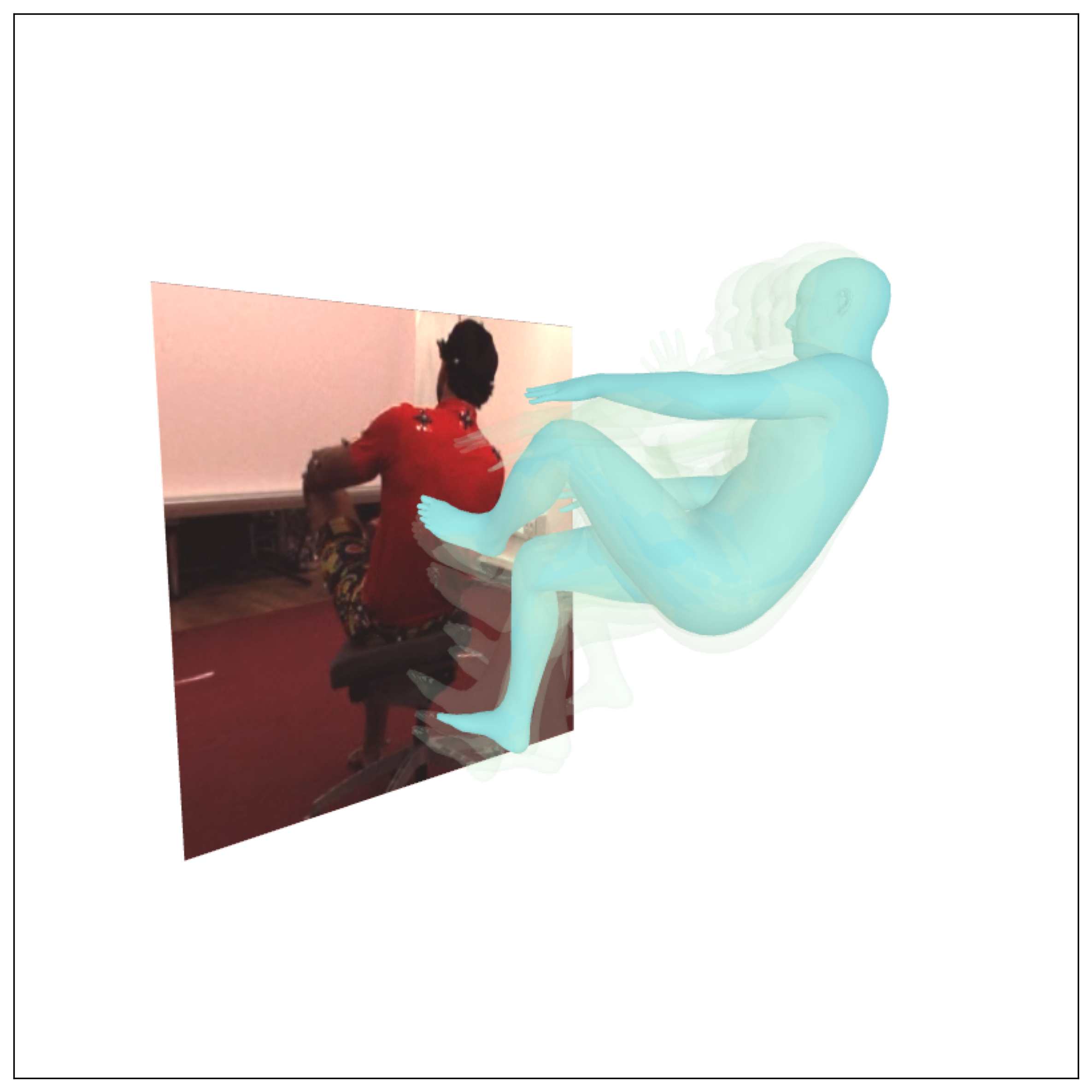}
    \end{subfigure}
    \begin{subfigure}[t]{0.24\textwidth}
        \centering
        \includegraphics[trim= 20 20 20 20, clip, width=\linewidth]{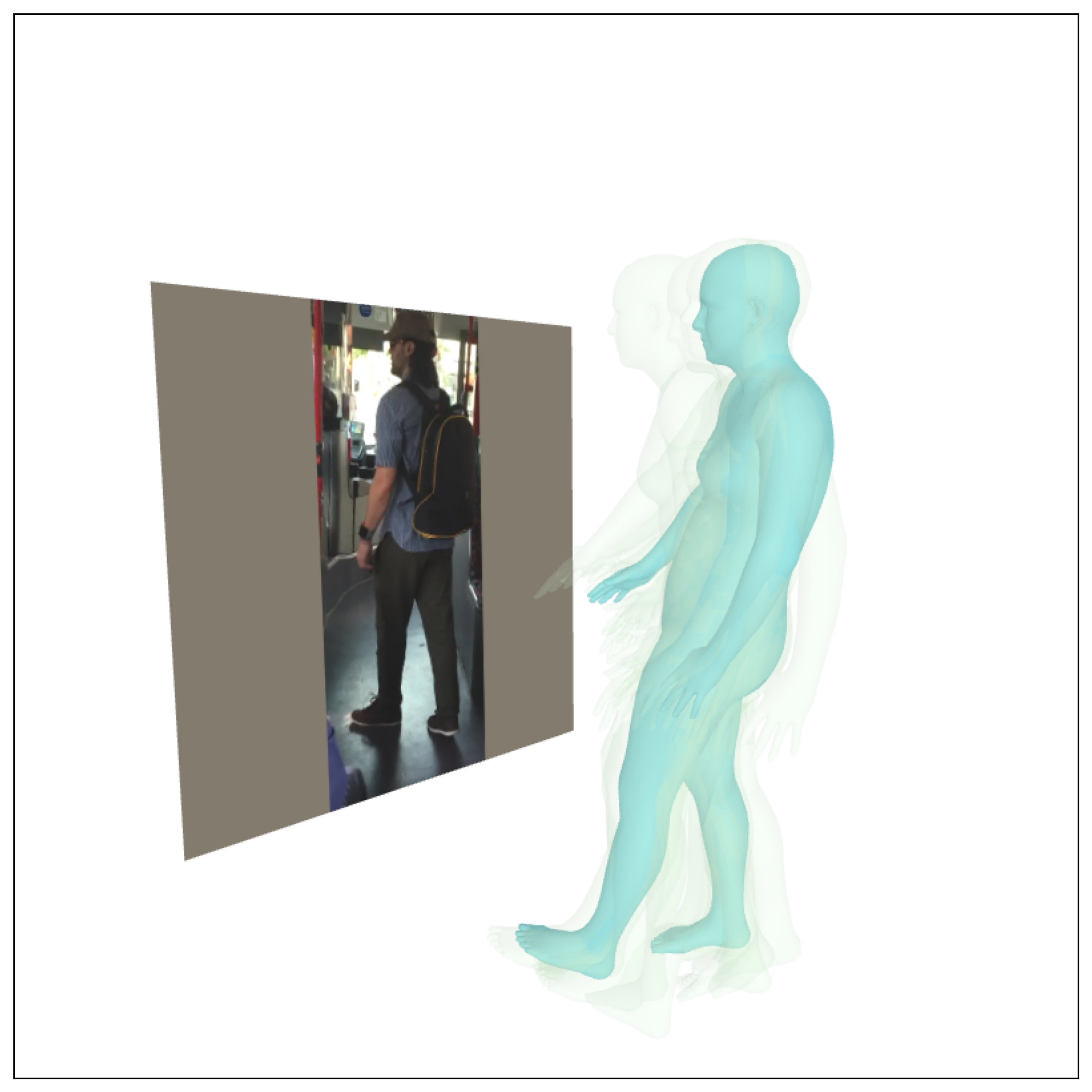}
    \end{subfigure}
    \begin{subfigure}[t]{0.24\textwidth}
        \centering
        \includegraphics[trim= 20 20 20 20, clip, width=\linewidth]{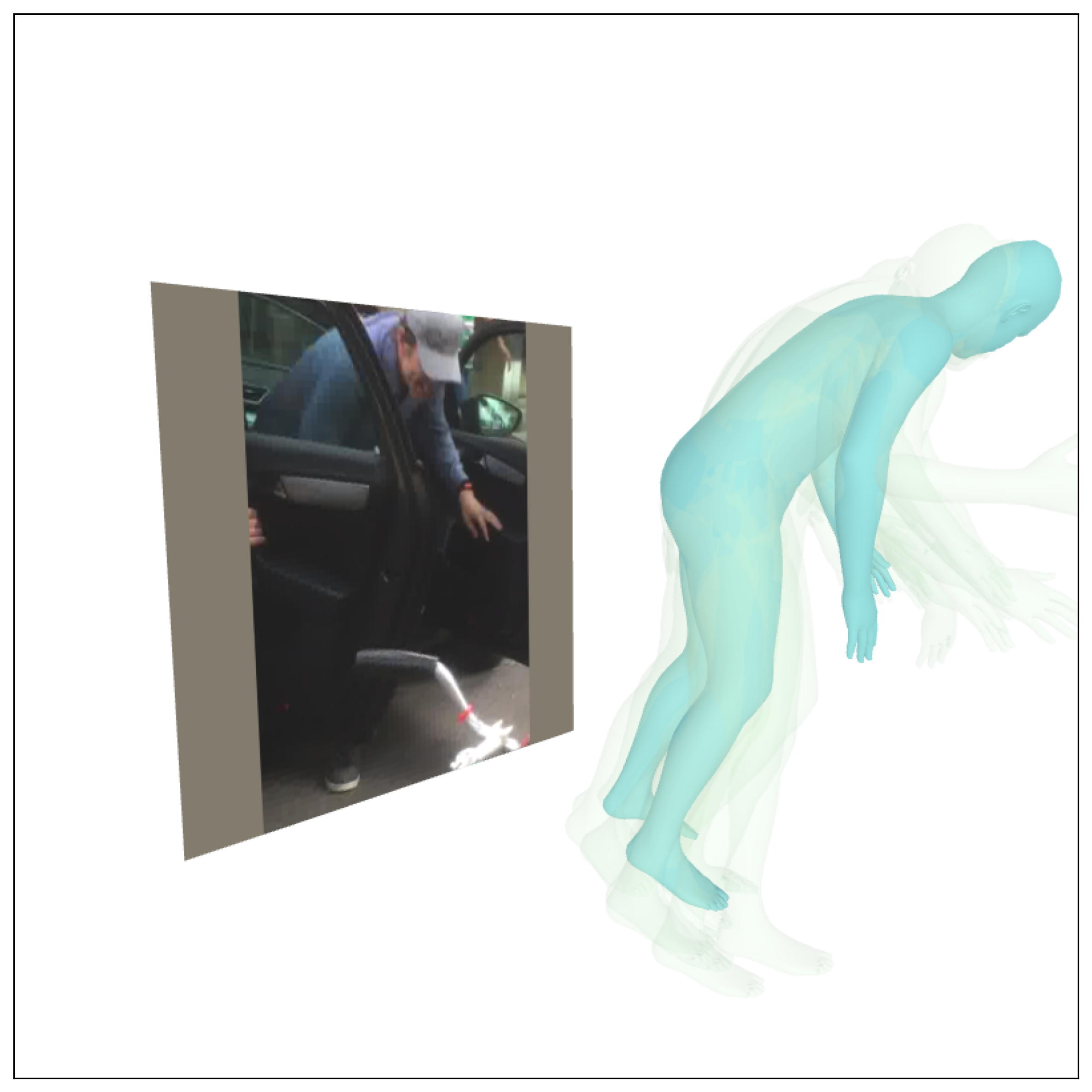}
    \end{subfigure}
    \begin{subfigure}[t]{0.24\textwidth}
        \centering
        \includegraphics[trim= 20 20 20 20, clip, width=\linewidth]{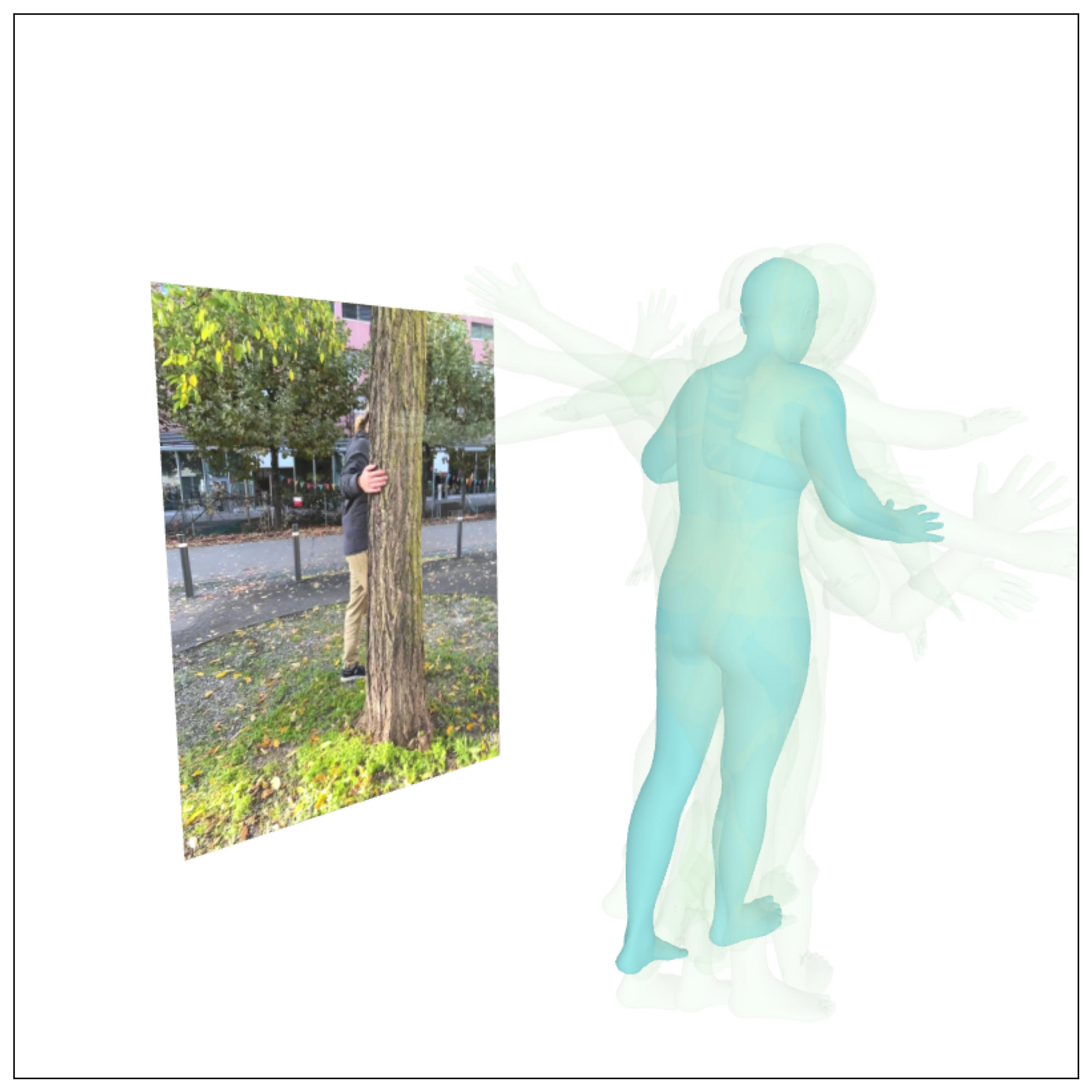}
    \end{subfigure}
    \caption{
    The qualitative results of CQF-HMR.
    We select some of the most challenging examples from the three datasets: Human3.6M (1\textsuperscript{st}), 3DPW (2\textsuperscript{nd} and 3\textsuperscript{rd}), and EMDB (4\textsuperscript{th}).
    The poses are visualized in camera coordinates with the view slightly rotated to the left for better visualization of the occluded body parts.
    The best hypothesis is shown in high contrast, whereas others are in lower intensity.
    }
    \label{fig:qualitative}
\end{figure*}

\textbf{On 3DPW}, we finetune the trained models on training split from~\cite{marcard_eccv18}, and we collect the evaluations on the training split in~\cref{tab:exp_3dpw}.
The results of the probabilistic methods are collected as the minimum errors out of $100$ sampled hypotheses.
In general, CQF-HMR performs on-par with the recent comparable baseline such as HuManiFlow or ScoreHypo.
Compared to the closest baseline ScoreHypo, the CQF-HMR~\textsubscript{($\mathcal{N}$)} has $2.3\%(1.5\unit{mm})$ of improvement in MPJPE, but worse PA-MPJPE with $1.5\%(0.6\unit{mm})$.
The CQF-HMR~\textsubscript{($\mathcal{D}$)} achieves worse results in both metrics.
This is due to the challenging examples of 3DPW that contain more occlusions and a higher level on uncertainty.
Sampling from the random Gaussian distribution helps with uncertainty coverage, \ie wider range of possible initial pose, compared to the VPoser prior that mostly produce up-right standing poses.
The method from~\citet{wehrbein_wacv25} is trained on a much larger database, the BEDLAM~\cite{black_bedlam_cvpr23} and AGORA~\cite{patel_agora_cvpr21} combined with the training set of 3DPW, thus the incomparable results.
Due to computational limitation, we cannot afford to train on the same scale and only keep their results in~\cref{tab:exp_3dpw} for referencing purposes.
In~\cref{tab:3dpw_div}, we additionally collect the Diversity metric that measures how spread the 3D joint hypotheses to their means.

\textbf{On EMDB}, we perform inference without extra fine-tuning on the subset proposed by~\cite{wehrbein_wacv25} and report the results in~\cref{tab:exp_emdb}.
We observe a similar performance comparison as with 3DPW.
The CQF-HMR~\textsubscript{($\mathcal{N}$)} achieves on-par results to the comparable baselines, \eg HuManiFlow~\cite{sengupta_cvpr23} or ScoreHypo~\cite{xu_cvpr24}, with sub $1\%$ difference on MPJPE and $3\%$ on PA-MPJPE between methods, showing the competitiveness of CQF-HMR on unseen data compared to related work.

\subsection{Comparison to different 3D rotations}
\label{subsec:exp_rotations}

We conduct an ablation study on the usage of different 3D rotation representations for the probabilistic human mesh recovery.
The experiments are evaluated on the Human3.6M's regular test set and the ambiguous set defined by~\cite{wehrbein_iccv21}.
The 3D rotation representations in this study are: the 3D Euler angles, the 3D axis-angle, and the 4D quaternion.
For quaternions, in addition to the OT mapping of \texttt{slerp}, we exanimate the performance with \texttt{lerp} and normalized \texttt{lerp}, \ie \texttt{nlerp}.
In~\cref{fig:rotations}, we visualize the integrated mapping of the 3D rotations with their respective OT flows.
It can be seen that Euler angles and axis-angle with \texttt{lerp} produce highly complex trajectories, while quaternion with \texttt{nlerp} produce smooth but not optimal mapping, \ie not reaching the target pose.
The proposed solution of quaternion with \texttt{slerp} produces the smoothest trajectory.

We collect the corresponding quantitative results of different 3D rotations in~\cref{tab:abl_rot3D}.
The advantage of quaternion with \texttt{slerp} OT flow is best described via the deterministic evaluations, where it outperforms other 3D rotations with the MPJPE improvement of $47.6\%$ and $8.5\%$ compared to axis-angle and Euler angles respectively when sampling from Gaussian ($\mathcal{N}$), or $33.9\%$ and $8.5\%$ when sampling from VPoser ($\mathcal{D}$).
The sub-par performance of 3D representations is due to their limitations during integration, \eg discontinuity or gimbal lock, preventing the flows reaching their target.
The probabilistic approach could help mitigate these issues by sampling a wider range of initial poses so that discontinuity or gimbal lock in 3D representation would not occur, but this approach contains randomness.
The quaternion with \texttt{slerp} and initial sampling from VPoser yields the strongest performance across all metrics.

\subsection{Qualitative results}
\label{subsec:exp_qualitative}

We present some qualitative results of CQF-HMR in~\cref{fig:qualitative}, with the challenging examples taken from the ambiguous split of Human3.6M, the 3DPW and the EMDB.
Note that all the 3D poses are visualized in camera coordinates, and the view is slightly rotated from the original camera view for better visualization of the estimated depth information.

\section{Conclusion}
\label{sec:conclusion}

In this project, we study the quaternion flow matching for the task of probabilistic 3D human mesh recovery using the optimal transport \texttt{slerp}.
Given heatmap inputs, we predict the smooth quaternion flows mapping a randomly sampled pose to a plausible hypothesis.
The experimental results show the advantages of quaternions with \texttt{slerp} over traditional 3D rotation representations in generating 3D human mesh hypotheses.
In comparison to related work, our CQF-HMR shows better performance on the Human3.6M dataset, while staying competitive to comparable methods in the more challenging settings of 3DWP and EMDB.

\textit{Limitations and future work}.
The quality of CQF-HMR is currently bounded by the performance of the 2D detector HRNet, especially on the ambiguity in human scale, \ie a far big person and a close small person have similar scaling.
Future work could extend CQF-HMR to consider the relationship between the human poses and the environment, \eg contacts or body-scene interactions, to extract more explicit information that help correcting the human scale, and disambiguating heavy-occluded scenarios.

\newpage
{
    \small
    \bibliographystyle{ieeenat_fullname}
    \bibliography{main}

@String{PAMI={IEEE Transactions on Pattern Analysis and Machine Intelligence}}

@String{CVPR={Proc. of Conference on Computer Vision and Pattern Recognition (CVPR)}}

@String{ECCV={Proc. of European Conference on Computer Vision (ECCV)}}

@String{ICCV={Proc. of International Conference on Computer Vision (ICCV)}}

@String{WACV={Proc. of the Winter Conference on Applications of Computer Vision (WACV)}}

@String(BMVC={Proc. of the British Machine Vision Conference (BMVC)})

@String{TOG={ACM Transactions on Graphics (TOG)}}

@String{NeurIPS={Advances in neural information processing systems}}

@String{ICLR={Proc. of International Conference on Learning Representations (ICLR)}}

@String{ICML={Proc. of the International Conference on Machine Learning (ICML)}}

@String(PAMI  = {IEEE TPAMI})

@String(CVPR  = {CVPR})

@String(ICCV  = {ICCV})

@String(ICCVW = {ICCVW})

@String(ECCV  = {ECCV})

@String(WACV  = {WACV})

@String(BMVC  =	{BMVC})

@String(TMLR  =	{TMLR})

@String(TOG   = {ACM TOG})

@String(NeurIPS  = {NeurIPS})

@String{ICLR = {ICLR}}

@String{ICML = {ICML}}

@article{mehta_vnect17,
    author    = {Mehta, Dushyant and Sridhar, Srinath and Sotnychenko, Oleksandr and Rhodin, Helge and Shafiei, Mohammad and Seidel, Hans‐Peter and Xu, Weipeng and Casas, Dan and Theobalt, Christian},
    title     = {{VNect:} Real‑time 3D Human Pose Estimation with a Single RGB Camera},
    journal   = TOG,
    volume    = {36},
    number    = {4},
    year      = {2017}}

@article{rogez_tpami19,
    author    = {Rogez, Gr{\'e}gory and Weinzaepfel, Philippe and Schmid, Cordelia},
    title     = {Localization‑Classification‑Regression for Whole‑Body Human Pose Estimation},
    journal   = PAMI,
    year      = {2019}}

@inproceedings{yeh_nips19,
    author    = {Yeh, Raymond A. and Hu, Yuan‑Ting and Schwing, Alexander G.},
    title     = {Chirality Nets for Human Pose Regression},
    booktitle = NeurIPS,
    volume    = {32},
    year      = {2019}}

@inproceedings{liu_cvpr20,
    author    = {Liu, Ruixu and Shen, Ju and Wang, He and Chen, Chen and Cheung, Sen-ching and Asari, Vijayan},
    title     = {Attention Mechanism Exploits Temporal Contexts: Real-Time 3D Human Pose Reconstruction},
    booktitle = CVPR,
    year      = {2020},
    pages     = {5064--5073}}

@inproceedings{sun_cvpr19,
    author    = {Sun, Ke and Xiao, Bin and Liu, Dong and Wang, Jingdong},
    title     = {Deep High-Resolution Representation Learning for Visual Recognition},
    booktitle = CVPR,
    year      = {2019}
    }

@inproceedings{martinez_iccv17,
    author    = {Martinez, Julieta and Hossain, Rayat and Romero, Javier and Little, James J.},
    title     = {A simple yet effective baseline for 3d human pose estimation},
    booktitle = ICCV,
    year      = {2017}
    }

@inproceedings{cai_iccv19,
    author    = {Cai, Yujun and Ge, Liuhao and Liu, Jun and Cai, Jianfei and Cham, Tat-Jen and Yuan, Junsong and Thalmann, Nadia Magnenat},
    title     = {Exploiting Spatial-Temporal Relationships for 3D Pose Estimation via Graph Convolutional Networks},
    booktitle = ICCV,
    year      = {2019}}

@inproceedings{pavllo_cvpr19,
    author    = {Pavllo, Dario and Feichtenhofer, Christoph and Grangier, David and Auli, Michael},
    title     = {3D human pose estimation in video with temporal convolutions and semi-supervised training},
    booktitle = CVPR,
    year      = {2019}
    }

@inproceedings{zheng_iccv21,
    author    = {Zheng, Ce and Zhu, Sijie and Mendieta, Matias and Yang, Taojiannan and Chen, Chen and Ding, Zhengming},
    title     = {3D Human Pose Estimation with Spatial and Temporal Transformers},
    booktitle = ICCV,
    year      = {2021}}

@inproceedings{zhao_cvpr23,
    author    = {Zhao, Qitao and Zheng, Ce and Liu, Mengyuan and Wang, Pichao and Chen, Chen},
    title     = {PoseFormerV2: Exploring Frequency Domain for Efficient and Robust 3D Human Pose Estimation},
    booktitle = CVPR,
    year      = {2023},
    pages     = {8877--8886}}

@inproceedings{peng_cvpr24,
    title     = {KTPFormer: Kinematics and Trajectory Prior Knowledge-Enhanced Transformer for 3D Human Pose Estimation},
    author    = {Peng, Jihua and Zhou, Yanghong and Mok, PY},
    booktitle = CVPR,
    pages     = {1123--1132},
    year      = {2024}}

@inproceedings{jahangiri_iccvw17,
    author    = {Jahangiri, Ehsan and Yuille, Alan L.},
    title     = {Generating Multiple Diverse Hypotheses for Human 3D Pose Consistent with 2D Joint Detections},
    booktitle = ICCVW,
    year      = {2017}}

@inproceedings{li_cvpr2022,
    title     = {MHFormer: Multi-Hypothesis Transformer for 3D Human Pose Estimation},
    author    = {Li, Wenhao and Liu, Hong and Tang, Hao and Wang, Pichao and Van Gool, Luc},
    booktitle = CVPR,
    pages     = {13147--13156},
    year      = {2022}}

@inproceedings{li_cvpr19,
    author    = {Li, Chen and Lee, Gim Hee},
    title     = {Generating Multiple Hypotheses for 3D Human Pose Estimation with Mixture Density Network},
    booktitle = CVPR,
    year      = {2019}
    }

@InProceedings{sharma_iccv19,
    author    = {Sharma, Saurabh and Varigonda, Pavan Teja and Bindal, Prashast and Sharma, Abhishek and Jain, Arjun},
    title     = {Monocular 3D Human Pose Estimation by Generation and Ordinal Ranking},
    booktitle = ICCV,
    year      = {2019}
    }

@inproceedings{wehrbein_iccv21,
    author    = {Wehrbein, Tom and Rudolph, Marco and Rosenhahn, Bodo and Wandt, Bastian},
    title     = {Probabilistic Monocular 3D Human Pose Estimation with Normalizing Flows},
    booktitle = ICCV,
    year      = {2021}
    }

@inproceedings{oikarinen_gmdn21,
    title     = {GraphMDN: Leveraging Graph Structure and Deep Learning to Solve Inverse Problems},
    author    = {Oikarinen, Tuomas P. and Hannah, Daniel C. and Kazerounian, Sohrob},
    booktitle = {International Joint Conference on Neural Networks (IJCNN)},
    year      = {2021},
    pages     = {1--9},
    publisher = {IEEE}
    }

@inproceedings{holmquist_iccv23,
    author    = {Holmquist, Karl and Wandt, Bastian},
    title     = {DiffPose: Multi-hypothesis Human Pose Estimation using Diffusion Models},
    booktitle = ICCV,
    year      = {2023}
    }

@inproceedings{rommel_nips24,
    title     = {ManiPose: Manifold-Constrained Multi-Hypothesis 3D Human Pose Estimation},
    author    = {Rommel, Cédric and Letzelter, Victor and Samet, Nermin and Marlet, Renaud and Cord, Matthieu and Pérez, Patrick and Valle, Eduardo},
    booktitle = NeurIPS,
    year      = {2024}
    }

@inproceedings{wang_cvpr26,
    author    = {Wang, Ti and Yu, Xiaohang and Mathis, Mackenzie Weygandt},
    title     = {FMPose3D: Monocular 3D Pose Estimation via Flow Matching},
    booktitle = CVPR,
    year      = 2026,
    }

@article{le_tmlr26,
    author    = {Le, Cuong and Melnyk, Pavlo and Wandt, Bastian and Wadenbäck, Mårten},
    title     = {Flow Matching for Probabilistic Monocular 3D Human Pose Estimation},
    journal   = TMLR,
    year      = {2026},
    }

@inproceedings{zanfir_eccv20,
    title     = {Weakly Supervised 3D Human Pose and Shape Reconstruction with Normalizing Flows},
    author    = {Zanfir, Andrei and Bazavan, Eduard Gabriel and Xu, Hongyi and Freeman, Bill and Sukthankar, Rahul and Sminchisescu, Cristian},
    booktitle = ECCV,
    year      = 2020}

@inproceedings{aliakbarian_cvpr22,
    author    = {Aliakbarian, Sadegh and Cameron, Pashmina and Bogo, Federica and Fitzgibbon, Andrew and Cashman, Thomas J.},
    title     = {FLAG: Flow-based 3D Avatar Generation from Sparse Observations},
    booktitle = CVPR,
    year      = 2022}

@inproceedings{sfikas_bmvcw25,
    author = {Sfikas, Giorgos and Nikolaidou, Konstantina and Papadopoulou, Foteini and Retsinas, George and Kesidis, Anastasios L.},
    title = {Are Euler angles a useful rotation parameterisation for pose estimation with Normalizing Flows?},
    booktitle = {BMVC Workshop},
    year = {2025}}

@inproceedings{kocabas_cvpr20,
    author    = {Kocabas, Muhammed and Athanasiou, Nikos and Black, Michael J.},
    title     = {VIBE: Video Inference for Human Body Pose and Shape Estimation},
    booktitle = CVPR,
    pages     = {5253--5263},
    year      = {2020}
    }

@inproceedings{choi_eccv20,  
    author    = {Choi, Hongsuk and Moon, Gyeongsik and Lee, Kyoung Mu},  
    title     = {Pose2Mesh: Graph Convolutional Network for 3D Human Pose and Mesh Recovery from a 2D Human Pose},  
    booktitle = ECCV,  
    year      = {2020}  
    }

@inproceedings{li_hybrik21,
    title     = {HybrIK: A Hybrid Analytical-Neural Inverse Kinematics Solution for 3D Human Pose and Shape Estimation},
    author    = {Li, Jiefeng and Xu, Chao and Chen, Zhicun and Bian, Siyuan and Yang, Lixin and Lu, Cewu},
    booktitle = CVPR,
    pages     = {3383--3393},
    year      = {2021}
    }

@inproceedings{li_cliff22,
    author    = {Li, Zhihao and Liu, Jianzhuang and Zhang, Zhensong and Xu, Songcen and Yan, Youliang},
    title     = {CLIFF: Carrying Location Information in Full Frames into Human Pose and Shape Estimation},
    booktitle = ECCV,
    pages     = {590--606},
    year      = {2022}
    }

@inproceedings{kolotouros_iccv19,
    author         = {Kolotouros, Nikos and Pavlakos, Georgios and Black, Michael J and Daniilidis, Kostas},
    title          = {Learning to Reconstruct 3D Human Pose and Shape via Model-fitting in the Loop},
    booktitle      = ICCV,
    year           = {2019}
    }

@inproceedings{goel_iccv23,
    author    = {Goel, Shubham and Pavlakos, Georgios and Rajasegaran, Jathushan and Kanazawa, Angjoo and Malik, Jitendra},
    title     = {Humans in 4D: Reconstructing and Tracking Humans with Transformers},
    booktitle = ICCV,
    year      = {2023},
    pages     = {14783--14794}
    }

@inproceedings{kolotouros_iccv21,
    author    = {Kolotouros, Nikos and Pavlakos, Georgios and Jayaraman, Dinesh and Daniilidis, Kostas},
    title     = {Probabilistic Modeling for Human Mesh Recovery},
    booktitle = ICCV,
    year      = {2021}
    }

@inproceedings{sengupta_iccv21,
    author    = {Sengupta, Akash and Budvytis, Ignas and Cipolla, Roberto},
    title     = {Hierarchical Kinematic Probability Distributions for 3D Human Shape and Pose Estimation from Images in the Wild},
    booktitle = ICCV,
    year      = {2021}}

@inproceedings{xu_cvpr24,
    author    = {Xu, Yuan and Ma, Xiaoxuan and Su, Jiajun and Zhu, Wentao and Yu, Qiao and Wang, Yizhou},
    title     = {ScoreHypo: Probabilistic Human Mesh Estimation with Hypothesis Scoring},
    booktitle = CVPR,
    month     = {June},
    year      = {2024}
}

@inproceedings{sengupta_cvpr23,
    author    = {Sengupta, Akash and Budvytis, Ignas and Cipolla, Roberto},
    title     = {HuManiFlow: Ancestor-Conditioned Normalising Flows on SO(3) Manifolds for Human Pose and Shape Distribution Estimation},
    booktitle = CVPR,
    year      = {2023}}

@inproceedings{wehrbein_wacv25,
    author    = {Wehrbein, Tom and Rudolph, Marco and Rosenhahn, Bodo and Wandt, Bastian},
    title     = {Utilizing Uncertainty in 2D Pose Detectors for Probabilistic 3D Human Mesh Recovery},
    booktitle = WACV,
    year      = {2025}}

@inproceedings{lipman_iclr23,
    title     = {Flow Matching for Generative Modeling},
    author    = {Lipman, Yaron  and Chen, Ricky T. Q. and Ben-Hamu, Heli and Nickel, Maximilian and Le, Matt },
    booktitle = ICLR,
    year      = {2023}
    }

@inproceedings{chen_nips18,
    author    = {Chen, Ricky T. Q. and Rubanova, Yulia and Bettencourt, Jesse and Duvenaud, David},
    title     = {Neural Ordinary Differential Equations},
    booktitle = NeurIPS,
    volume    = {31},
    pages     = {6571--6583},
    year      = {2018}
    }

@article{elfwing_nn18,
    author    = {Elfwing, Stefan and Uchibe, Eiji and Kenji Doya},
    title     = {Sigmoid-weighted linear units for neural network function approximation in reinforcement learning},
    journal   = {Neural Networks},
    volume    = {107},
    pages     = {3--11},
    year      = {2018},
    }

@article{loper_acm15,
    author    = {Loper, Matthew and Mahmood, Naureen and Romero, Javier and Pons‑Moll, Gerard and Black, Michael J.},
    title     = {{SMPL}: A Skinned Multi-Person Linear Model},
    journal   = TOG,
    volume    = {34},
    number    = {6},
    pages     = {248:1--248:16},
    year      = {2015}
    }

@inproceedings{pavlakos_cvpr19,
    author    = {Pavlakos, Georgios and Choutas, Vasileios and Ghorbani, Nima and Bolkart, Timo and Osman, Ahmed A. A. and Tzionas, Dimitrios and Black, Michael J.},
    title     = {Expressive Body Capture: 3D Hands, Face, and Body from a Single Image},
    booktitle = CVPR,
    year      = {2019}}

@inproceedings{le_iclr26,
    title     = {QuaMo: Quaternion Motions for Vision-based 3D Human Kinematics Capture},
    author    = {Le, Cuong and Melnyk, Pavlo and Waldmann, Urs and Wadenbäck, Mårten and Wandt, Bastian},
    booktitle = ICLR,
    year      = {2026},
    url       = {https://openreview.net/forum?id=em0jPLYjaS}}

@book{yang_spacecraft19,
    author    = {Yang, Yaguang},
    title     = {Spacecraft modeling, attitude determination, and control: quaternion-based approach},
    publisher = {CRC Press},
    year      = {2019}}

@book{kuipers_quaternions99,
    author    = {Kuipers, Jack B},
    title     = {Quaternions and rotation sequences: a primer with applications to orbits, aerospace, and virtual reality},
    publisher = {Princeton university press},
    year      = {1999}}

@article{golabek_aircraft22,
    author    = {Gołąbek, Michał and Welcer, Michał and Szczepański, Cezary and Krawczyk, Mariusz and Zajdel, Albert and Borodacz, Krystian},
    title     = {Quaternion Attitude Control System of Highly Maneuverable Aircraft},
    journal   = {Electronics},
    volume    = {11},
    year      = {2022},
    number    = {22},
    article-number = {3775},
    ISSN      = {2079-9292},
    DOI       = {10.3390/electronics11223775}}

@book{vuik_numericalode23,
    author    = {Vuik, Cornelis and Vermolen, F.J. and van Gijzen, M.B. and Vuik, Thea},
    title     = {Numerical Methods for Ordinary Differential Equations},
    publisher = {TU Delft OPEN Publishing},
    year      = {2023}}

@article{ionescu_pami14,
    author    = {Ionescu, Catalin and Papava, Dragos and Olaru, Vlad and Sminchisescu, Cristian},
    title     = {Human3.6M: Large Scale Datasets and Predictive Methods for 3D Human Sensing in Natural Environments}, 
    journal   = PAMI, 
    volume    = {36},
    number    = {7},
    pages     = {1325--1339},
    year      = {2014}}

@inproceedings{marcard_eccv18,
    author    = {Marcard, Timo von and Henschel, Roberto and Black, Michael J. and Rosenhahn, Bodo and Pons-Moll, Gerard},
    title     = {Recovering Accurate 3D Human Pose in The Wild Using IMUs and a Moving Camera},
    booktitle = ECCV,
    year      = {2018}}

@inproceedings{kaufmann_iccv23,
    author    = {Kaufmann, Manuel and Song, Jie and Guo, Chen and Shen, Kaiyue and Jiang, Tianjian and Tang, Chengcheng and Z{\'a}rate, Juan Jos{\'e} and Hilliges, Otmar},
    title     = {{EMDB}: The {E}lectromagnetic {D}atabase of {G}lobal 3{D} {H}uman {P}ose and {S}hape in the {W}ild},
    booktitle = ICCV,
    year      = {2023}}

@inproceedings{patel_agora_cvpr21,
    author    = {Patel, Priyanka and Huang, Chun-Hao P. and Tesch, Joachim and Hoffmann, David T. and Tripathi, Shashank and Black, Michael J.}, 
    title     = {{AGORA}: Avatars in Geography Optimized for Regression Analysis}, 
    booktitle = CVPR, 
    month     = jun,
    year      = {2021},
    month_numeric = {6}}

@inproceedings{black_bedlam_cvpr23,
    author    = {Black, Michael J. and Patel, Priyanka and Tesch, Joachim and Yang, Jinlong}, 
    title     = {{BEDLAM}: A Synthetic Dataset of Bodies Exhibiting Detailed Lifelike Animated Motion},
    booktitle = CVPR,
    pages     = {8726-8737},
    month     = jun,
    year      = {2023},
    month_numeric = {6}}

@inproceedings{rempe_humor21,
    author    = {Rempe, Davis and Birdal, Tolga and Hertzmann, Aaron and Yang, Jimei and Sridhar, Srinath and Guibas, Leonidas J.},
    title     = {HuMoR: 3D Human Motion Model for Robust Pose Estimation},
    booktitle = ICCV,
    year      = {2021}}

@inproceedings{zhao_cvpr26,
    author    = {Zhao, Yiwen and Zheng, Ce and Wang, Yufu and Yang, Hsueh-Han Daniel and Wen, Liting and Jeni. Laszlo A.},
    title     = {OnlineHMR: Video-based Online World-Grounded Human Mesh Recovery},
    booktitle = CVPR,
    year      = 2026}

@inproceedings{bogo_eccv16,
    author    = {Bogo, Federica and Kanazawa, Angjoo and Lassner, Christoph and Gehler, Peter and Romero, Javier and Black, Michael J.},
    title     = {Keep it SMPL: Automatic Estimation of 3D Human Pose and Shape from a Single Image},
    booktitle = ECCV,
    year      = 2016}

@inproceedings{yue_icml25,
    title     = {ReQFlow: Rectified Quaternion Flow for Efficient and High-Quality Protein Backbone Generation},
    author    = {Yue, Angxiao and Wang, Zichong and Xu, Hongteng},
    booktitle = ICML,
    year      = {2025}}
}
\newpage
\clearpage
\setcounter{page}{1}
\setcounter{section}{5}
\setcounter{table}{6}
\maketitlesupplementary

\section{Additional experiments}
\label{suppsec:add_exps}

\subsection{Lifting condition}
\label{suppsubsec:condition}

We additionally conduct an ablation in~\cref{tab:supp_condition} for verifying the contribution of the selected conditioning network GCN.
The architectures in comparison are multi-layer perceptron (MLP) and the transformer.
We replace the GCN module in CQF-HMR with other architectures and keeping the number of training parameters comparable (Param.), approximately $4.6\textup{M} \rightarrow 4.9\textup{M}$.
We report the processing time (Time), the minimum MPJPE and PA-MPJPE on the regular test set of the Human3.6M dataset, from a set of $100$ hypotheses.
The GCN achieves the best trade-off between complexity and pose prediction error.
The transformer has slightly lower MPJPE ($0.3$\unit{\milli\metre}, $0.6\%$) in the sampling from Gaussian ($\mathcal{N}$) setting but has almost $\times 2$ processing time for generating $100$ hypotheses.
The GCN with sampling from the prior VPoser ($\mathcal{D}$) has the best results.

\begin{table}[ht]
    \centering
    \small
    \setlength\tabcolsep{1pt}
    \begin{tabularx}{\linewidth}{@{\extracolsep{\fill}}lccccc}
        \toprule
        Condition & Initial & Param. & Time & MPJPE & PA-MPJPE \\
        \midrule
        MLP & $\mathcal{N}$ & $4.70$M & $236.54$\unit{\milli\second} & $44.1{\scriptstyle\mathrm{\pm0.1}}$ & $30.8{\scriptstyle\mathrm{\pm0.1}}$ \\
        Transformer & $\mathcal{N}$ & $4.89$M & $302.62$\unit{\milli\second} & $43.2{\scriptstyle\mathrm{\pm0.1}}$ & $30.1{\scriptstyle\mathrm{\pm0.2}}$ \\
        GCN & $\mathcal{N}$ & $4.65$M & $150.08$\unit{\milli\second} & $43.5{\scriptstyle\mathrm{\pm0.3}}$ & $30.2{\scriptstyle\mathrm{\pm0.1}}$ \\
        \midrule
        MLP & $\mathcal{D}$ & $4.70$M & $237.21$\unit{\milli\second} & $41.9{\scriptstyle\mathrm{\pm0.1}}$ & $30.0{\scriptstyle\mathrm{\pm0.2}}$\\
        Transformer & $\mathcal{D}$ & $4.89$M & $304.72$\unit{\milli\second} & $41.7{\scriptstyle\mathrm{\pm0.3}}$ & $29.9{\scriptstyle\mathrm{\pm0.3}}$ \\
        GCN & $\mathcal{D}$ & $4.65$M & $155.82$\unit{\milli\second} & $41.3{\scriptstyle\mathrm{\pm0.2}}$ & $29.3{\scriptstyle\mathrm{\pm0.2}}$ \\
        \bottomrule
    \end{tabularx}
    \caption{Quantitative results of different learning architectures for the conditioning module.}
    \label{tab:supp_condition}
\end{table}

\subsection{Computational complexity}
\label{suppsubsec:complexity}

We provide the computational complexity and inference speed comparison between different 3D representations in~\cref{tab:supp_complexity}.
The inference speeds are averaged over $1000$ runs on the NVIDIA A100 GPU device.
From the Tab.~5 of the main paper, we record that using quaternion with \texttt{slerp} achieves the lowest pose error.
The trade-off in speed is approximately $10-20$\unit{\milli\second} longer than the linear quaternion (\texttt{lerp}) or other common 3D representations, \eg axis-angle, Euler angles.
The bottleneck is mainly due to the conversions back-and-fourth between the SMPL's original representation axis-angle to quaternion.
This is an open challenge and we are actively working on a more optimized implementation.

\begin{table}[ht]
    \centering
    \small
    \setlength\tabcolsep{1pt}
    \begin{tabularx}{\linewidth}{@{\extracolsep{\fill}}lcccc}
        \toprule
        Rotation & Initial & Param. & $1$ hypo & $100$ hypos \\
        \midrule
        Axis-angle & $\mathcal{N}$ & $4.60$M & $38.38$\unit{\milli\second} & $134.46$\unit{\milli\second} \\
        Euler angles & $\mathcal{N}$ & $4.60$M & $40.49$\unit{\milli\second} & $135.87$\unit{\milli\second} \\
        Quaternion~(\texttt{lerp}) & $\mathcal{N}$ & $4.65$M & $38.28$\unit{\milli\second} & $130.23$\unit{\milli\second} \\
        Quaternion~(\texttt{slerp}) & $\mathcal{N}$ & $4.65$M & $55.29$\unit{\milli\second} & $150.08$\unit{\milli\second} \\
        \midrule
        Axis-angle & $\mathcal{D}$ & $4.60$M & $40.03$\unit{\milli\second} & $135.56$\unit{\milli\second} \\
        Euler angles & $\mathcal{D}$ & $4.60$M & $41.79$\unit{\milli\second} & $135.25$\unit{\milli\second} \\
        Quaternion~(\texttt{lerp}) & $\mathcal{D}$ & $4.65$M & $40.49$\unit{\milli\second} & $136.86$\unit{\milli\second} \\
        Quaternion~(\texttt{slerp}) & $\mathcal{D}$ & $4.65$M & $57.65$\unit{\milli\second} & $155.82$\unit{\milli\second} \\
        \bottomrule
    \end{tabularx}
    \caption{
    Complexity comparison of human pose hypothesis generation between different 3D representations on a A100 device.
    Hypo means hypothesis.}
    \label{tab:supp_complexity}
\end{table}



\end{document}